\documentclass[review]{elsarticle}

\usepackage[margin=1in, top=1in, bottom=1in]{geometry}

\usepackage{hyperref}
\usepackage{booktabs}
\usepackage{float}
\usepackage{comment}
\usepackage{subfigure}
\usepackage{adjustbox}
\usepackage{graphicx}
\usepackage{algpseudocode}
\usepackage[flushleft]{threeparttable}
\usepackage{amsmath}
\usepackage{amssymb}
\usepackage{pgfplots}
\usepackage{algorithm}
\usepackage{algpseudocode}
\usepackage{xfrac}
\usepackage{multirow}
\usepackage{figsize}
\usepackage{caption}
\usepackage{array}
\usepackage{siunitx}
\usepackage{tikz}
\usepackage{collcell}
\usepackage{pifont}

\usepackage{pgfplotstable}
\usepackage{colortbl}

\pgfplotsset{
    colormap={confusionmatrix}{
        color(0cm)=(white);
        color(1cm)=(green!75!black)
    }
}

\let\svtikzpicture\tikzpicture
\def\tikzpicture{\noindent\svtikzpicture}
\usepackage{pgfplots}
\usepgfplotslibrary{external} 
\def\hlinewd#1{%
\noalign{\ifnum0=`}\fi\hrule \@height #1 %
\futurelet\reserved@a\@xhline} 
\pgfplotsset{compat=newest} 
\usepgfplotslibrary{units} 
\usepgfplotslibrary{groupplots}

\journal{arXiv}

\begin{document}

\begin{frontmatter}

\title{DeepForge: Leveraging AI for Microstructural Control in Metal Forming via Model Predictive Control}
\author[ETH]{Jan Petrik\corref{mycorrespondingauthor}}
\cortext[mycorrespondingauthor]{Corresponding author. E-mail address: jpetrik@ethz.ch}
\author[ETH]{Markus Bambach}
\address[ETH]{Advanced Manufacturing Laboratory, ETH Zurich, Switzerland}

\begin{abstract}
This study presents a novel method for microstructure control in closed die hot forging that combines Model Predictive Control (MPC) with a developed machine learning model called DeepForge. DeepForge uses an architecture that combines 1D convolutional neural networks and gated recurrent units. It uses surface temperature measurements of a workpiece as input to predict microstructure changes during forging. The paper also details DeepForge's architecture and the finite element simulation model used to generate the data set, using a three-stroke forging process. The results demonstrate DeepForge's ability to predict microstructure with a mean absolute error of 0.4$\pm$0.3\,\%. In addition, the study explores the use of MPC to adjust inter-stroke wait times, effectively counteracting temperature disturbances to achieve a target grain size of less than 35 microns within a specific 2D region of the workpiece. These results are then verified experimentally, demonstrating a significant step towards improved control and quality in forging processes where temperature can be used as an additional degree of freedom in the process.
\end{abstract}

\begin{keyword}
Hot Closed-Die Forging; Model Predictive Control; Machine Learning; Microstructure Prediction; Finite Element Method\end{keyword}

\end{frontmatter}

\section{Introduction}
\label{sec:introduction}

Hot closed-die forging is a manufacturing process widely used in industries requiring high performance components, particularly in the automotive and aerospace sectors. The process begins by heating a billet, a block of metal, to temperatures above its recrystallisation point. This heating prepares the metal for forming. Once the billet has reached the desired temperature, it is subjected to compressive forces within a die, a tool designed to form the material into the desired shape. A significant advantage of closed-die forging at such elevated temperatures is the reduced force required to form the workpiece. The high temperatures not only make the metal more ductile, but also facilitate microstructural changes within the material. \cite{kevorkijan2003az80, tetsui2003strengthening}  \\

Despite its advantages, the closed-die forging process is subject to a wide range of variability that can significantly affect the microstructure and consequently the mechanical properties of the final product. Variations in billet microstructure, heating temperatures, handling times and die interactions add to the complexity of achieving consistent and optimum product quality even in the presence of process disturbances. These challenges highlight the need for realistic control mechanisms using fast models within the forging process to ensure the desired results, particularly in terms of microstructural properties that directly influence the strength, toughness and fatigue resistance of forged components. \\

The state-of-the-art in closed-die forging can be organized into several key categories: microstructure and mechanical properties, process control in forging, measurement of microstructure, and machine learning in forging processes. The current literature in each of these categories is described in a paragraph on the following pages. \\


As far as the microstructure and mechanical properties are concerned, they have a significant influence on its strength, toughness and fatigue resistance. For example, in forged steel gears, austenite grain size is known to have a significant effect on properties after final heat treatment. The influence of grain size on tooth root strength was investigated by \citet{janbein2015einfluss}, who showed that fine austenite grain size increases tooth root strength. In the research conducted by \citet{anzinger1991werkstoff}, it was found that reducing the grain size parameter G in 16MnCr5 steel (as specified in DIN 50106) from 11-12 to 7-8 could result in an improvement in root fatigue strength of approximately 12\,\%. In addition, \citet{weck1992einfluss} investigated the effect of material composition, heat treatment and manufacturing processes on the load carrying capacity of the tooth root, establishing a relationship between the amount of retained austenite and the strength of the tooth root. The grain size achieved in the austenitic state is a direct result of the thermal and mechanical history that the material undergoes during the forging process and is reflected in the properties of the final product, even if heat treatment is applied after forging. Precise control of the microstructure evolution during forging is therefore essential to achieve the desired mechanical properties in the final part. 

Moreover, differences in the initial microstructure of billets, whether between or within batches, deviations in billet temperature after heating, inconsistencies in transport times from furnace to press, and fluctuations in friction and heat transfer between the workpiece and dies all contribute to the final microstructure after forging. \\


Process control in forging was explored by \citet{allwood2016closed}, who investigated the potential of advances in property prediction to improve product quality. They highlight a growing trend in the development of closed-loop control systems in metal forming. However, their survey suggests that current research is predominantly focused on controlling product geometry and preventing defects, with comparatively less attention paid to the critical aspects of temperature and microstructure control. By providing a framework for analysing and developing future control systems, the authors encourage research and growth in these less explored areas and demonstrate the need for comprehensive control approaches in metal forming processes.

Furthermore, closed-loop control approaches could be indeed effectively utilized in closed-die forging to manage disturbances that are challenging to predict and incorporate into process design. \citet{stebner2023monitoring} have reviewed control concepts in forming that aim to decouple the evolution of workpiece geometry from its microstructure and properties across various forming processes, highlighting different levels of control flexibility. This research underscores that successful closed-loop control necessitates sufficient degrees of freedom to manage target properties without compromising the desired workpiece geometry. In closed-die forging, where options to independently control geometry and microstructure are limited, surrogate variables like ram speed are often used for control.

Meanwhile, \citet{lin2018precise} developed a One-Step-Ahead Model Predictive Control strategy using a backpropagation approach, specifically tailored for forging processes. This strategy employs two online-capable neural networks – a predictive neural network and a control neural network – to manage the hydraulic press in closed-die forging, with a particular focus on predicting ram velocity. However, it is crucial to note that this approach does not directly target the control of workpiece properties. \\


What makes the closed-loop control approaches in the hot closed-die forging process difficult, in addition to the limited degrees of freedom, is the limited accessibility. This contrasts with open-die forging, where \citet{rosenstock2014online} implemented the "LaCam (R) FORGE" Time-of-Flight measurement system. This system estimates and visualizes the temperature and microstructure of the workpiece based on surface temperature measurements, providing plant operators with a visual feedback mechanism to adjust the forging process accordingly. However, the concept of automated closed-loop control was not explored in this context. \citet{levesque2006thickness} also made strides in monitoring wall thickness and grain size in rolled seamless pipes using laser ultrasound technology. Yet, the integration of such advanced measurement systems into the closed-die forging process is significantly more complex. The challenges stem from the high-speed nature of the process and the difficulty in making contact with a lubricated workpiece.

Furthermore, the issues highlighted in closed-die forging, where it is difficult to measure the internal state of the workpiece, become even more apparent when the gap between what needs to be predicted and what can be measured directly is considered. This difficulty is primarily due to the complexity of directly measuring microstructural characteristics during the forming process. As a result, estimating the microstructural state and the resultant properties for property control necessitates the use of material and process models. These models, often mean field material models, are crucial for simulating microstructure evolution, describing the average changes in state variables such as dislocation densities and grain sizes. Early foundational work in this area, such as the phenomenological models by \citet{sellars1986modelling}, employed algebraic equations to describe the evolution of flow stress and grain size due to recrystallization processes in hot working.

However, the inherent uncertainties in both the forging process conditions and these material models impose limitations on their application, even in the design and feedforward control of forging processes. \citet{henke2014optimization} demonstrated this by quantifying the uncertainty in a microstructure model for hot forging of steel, using a gear wheel as an example. Their findings revealed a "fat-tail" probability distribution in the resultant grain size, indicating the occurrence of outliers with a finite probability. To address this, they proposed a resampling method for uncertainty quantification, which requires each lab-scale experiment used to parameterize the material model to be repeated at least three times. While this approach emphasizes the importance of robust process design, it also highlights the practical challenges and high costs associated with uncertainty quantification in forging.

The limited ability to observe the microstructure directly is a key reason why process control in forging tends to focus on other targets, such as workpiece geometry. In open-die forging processes, for instance, \citet{nye2001real} implemented feedback control strategies that emphasized controlling the workpiece geometry, explicitly excluding microstructure evolution. 

Building on the challenges of controlling microstructure in forging, there is a growing emphasis on developing process models that can operate online. \citet{homberg2023softsensors} introduced the concept of "soft sensors", differentiating them from state observers. They showcased various contemporary applications of soft sensors in forming technology. Essentially, soft sensors combine hardware sensors with process models to infer non-observable quantities, such as microstructural features, from measurable quantities.

Expanding on this concept, \citet{bambach2015instabilities} presented an innovative method for controlling microstructure and properties during drop forging. The method utilizes impact energy and pause time as primary control variables. Rather than tracking microstructure evolution at each point within the workpiece, the method focuses on specific regions for desired microstructural states and monitors the progression of these regions' boundaries. A fast, data-driven surrogate model is proposed to predict the boundary location based on control input, effectively acting as a soft sensor that estimates the microstructural state using process information. Although the soft sensor has been experimentally validated, it has not yet been integrated into a control framework. The challenge lies in the offline calibration of these models, which may not provide the necessary robustness during actual forging processes. To address this, \citet{bambach2018simulation} suggests a framework that combines a data-driven microstructure model with a particle filter to update predictions with new measurements. \\


Furthermore, there is a growing research interest in the application of AI in material plasticity and temperature modeling. \citet{huang2020machine} introduced a machine learning-based plasticity model using proper orthogonal decomposition, aimed at predicting material behavior under plastic deformation. Adding to this, \citet{ibragimova2022convolutional} developed a framework using convolutional neural networks for crystal plasticity finite element analysis, focusing on predicting localized deformation in metals. \citet{petrik2023crystalmind} went further by introducing an AI-based algorithm, CrystalMind, which predicts recrystallization and deformation in a three-dimensional framework within the forging process.

In parallel, advancements have been made in the prediction of transient temperatures using AI. \citet{sarkar2021machine} applied machine learning to predict and analyze transient temperatures in submerged arc welding. \citet{kumar2023physics} developed physics-informed machine learning models for predicting transient temperature distribution in ferritic steel during directed energy deposition. \citet{xie2022two} created a two-dimensional transient heat transfer model for moving quenching jets, based on machine learning techniques.

The combination of AI with plasticity and temperature modeling was further explored by \citet{peng2020coupling}, who focused on coupling physics with machine learning to predict properties of high-temperature alloys. \citet{wen2021physics} developed a physics-driven machine learning model for temperature and time-dependent deformation in lithium metal, including its finite element implementation.

Moreover, in the microstructure domain, \citet{ford2021machine} explored the use of machine learning for accelerated property prediction of two-phase materials, integrating microstructural descriptors with finite element analysis. \citet{yang2021self} proposed a self-supervised learning approach for predicting microstructure evolution, utilizing convolutional recurrent neural networks. \citet{sengodan2021prediction} also contributed by predicting two-phase composite microstructure properties through machine learning of reduced dimensional structure-response data. \\

As previously highlighted, the precise prediction and optimisation of microstructural variables, such as grain size, is paramount in achieving the desired mechanical properties and ensuring the integrity of the final forged components. Despite the advances in AI for material plasticity, temperature modelling and microstructure prediction, there is a significant gap in the application of these technologies in the forging industry. In particular, there is a notable lack of sophisticated soft sensors capable of accurately predicting the full-scale thermomechanical solution, including microstructural variables, based on quantifiable inputs in real forging scenarios. In addition, the implementation of closed-loop control systems for microstructure regulation in workpieces during hot forging processes remains unachieved. \\

To address the challenges and research gaps mentioned above, this study aims to provide answers to the following research questions:

\begin{itemize}
    \item What type of machine learning algorithm and architecture is best suited to act as a soft sensor, i.e. can the full field output be predicted if the network is fed only partial information, e.g. surface temperatures, rather than the full temperature field?
    \item Can such an approach be integrated into a control framework to achieve a desired part microstructure with the potential to be used in real time during the process?
    \item How can temperature variations be effectively utilised in the forging process if they are treated as additional degrees of freedom in the forging process?
\end{itemize}

Answering the research questions outlined above should enable the development of a machine learning architecture that can accurately and robustly predict microstructural outcomes in forging processes. This architecture would use partial information, such as surface temperatures, to predict full-field outcomes, effectively acting as a soft sensor within a real-time control framework. Integrating this predictive capability into forging process control would allow dynamic adjustment of process parameters to achieve desired microstructures. By treating temperature variations as controllable parameters, the forging process can be optimised for improved material properties, demonstrating the potential of machine learning to improve manufacturing precision and efficiency.\\

Section \ref{sec:mathandmethods} presents a brief overview of the forging process and its FE model, as well as the experimental setup and data generation with its pre-processing. In addition, this section outlines the workflow, architecture and training settings of DeepForge, together with a description of the Model Predictive Control (MPC) setup. Section \ref{sec:results} reports both qualitative and quantitative results obtained by the trained DeepForge, as well as a comparison of the results generated by DeepForge with experimental results. Furthermore, section \ref{sec:discussion} discusses the research questions as well as the comparison with the state-of-the-art. Finally, Section \ref{sec:summ_conc} provides a brief summary of the work and highlights the main contributions.

\section{Materials and Methods}
\label{sec:mathandmethods}

\subsection{Forging Process and FE Model}
\label{sec:hotworkingprocess}
\begin{figure}[H]
   \centering
   \includegraphics[width=1.0\textwidth]{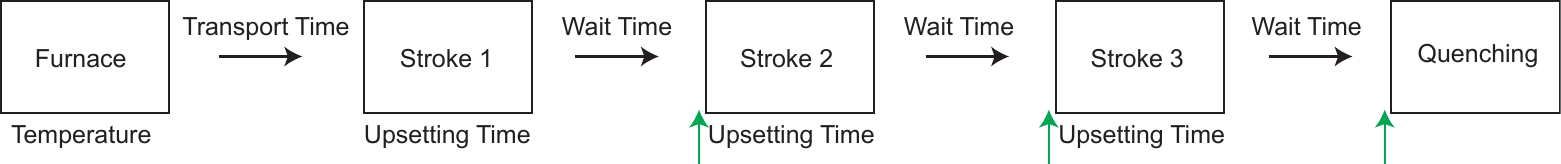}
   \caption{A diagram illustrating the simulation that involves a three-stage hot closed-die forging process, in which a specimen is moved from a preheated oven to a designated \textit{temperature} in \textit{transport time} and then subjected to three forging strokes. Each forging stroke's time is indicated by the \textit{upsetting time}, while the interval between two successive strokes is expressed by a parameter called \textit{wait time}. Finally, the last \textit{wait time} is between the final stroke and the quenching. It is also worth noting that the green arrows visualise the moments when the simulation outputs the microstructure of the workpiece.}
   \label{fig:deepforgefemschema}
\end{figure}

This work draws upon a three-stage upsetting process in order to investigate the above mentioned research questions. The workflow of the process is visualised in Figure \ref{fig:deepforgefemschema}.  

To train, validate and test a machine-learning based algorithm called DeepForge, a finite element (FE) model of this process was used.  The FE model was developed in Abaqus/Standard and involves the simulation of a three-phase hot forming process, integrating both thermal and mechanical analysis. In addition, MATLAB is utilized after each phase of the simulation to implement static recrystallization and grain growth models. This approach is crucial for pinpointing regions where the grain size is less than 35 micrometers (see \citet{bambach2021soft} for more details). \\

\begin{figure}[H]
   \centering
   \includegraphics[width=1.0\textwidth]{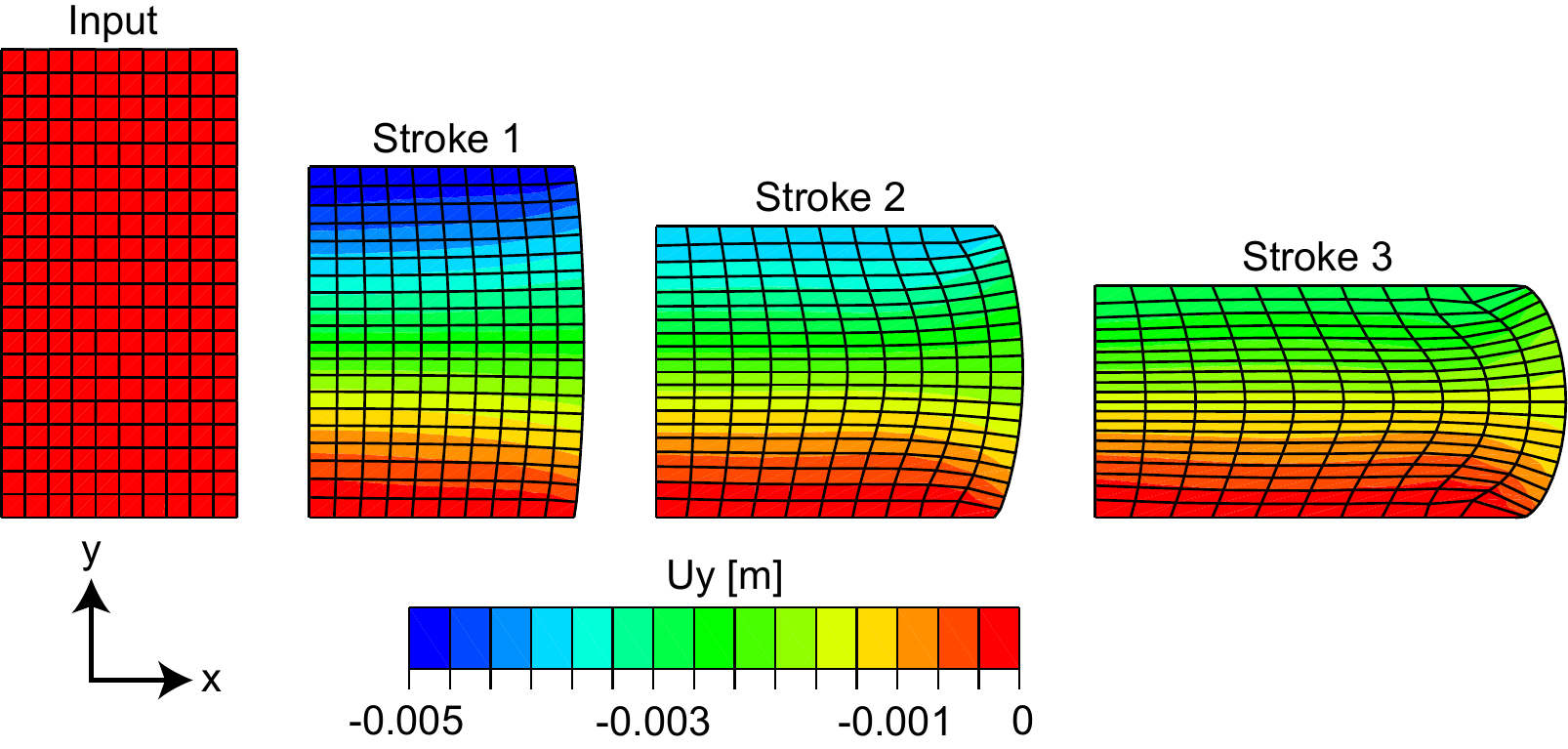}
   \caption{Visualization of the three-stroke hot forging FE-based simulation, where the colour bar corresponds to displacement in y denoted as $U_{y}$. The specimen consists of 20$\times$10 finite elements giving a total of 231 nodal values. It should also be noted that, due to axial symmetry, only one-half of the part is simulated.}
   \label{fig:deepforgesim}
\end{figure}

Figure \ref{fig:deepforgesim} provides a visualization of the FE-based simulation for the three-stroke forging process, illustrating the progressive deformation of the 2D workpiece that represents a mid-section of a cylinder. This specific geometry was chosen because it is commonly used in hot forging for semi-finished products, i.e. billets. \cite{dudra1990investigation, kim2001analysis}

\subsection{Experimental Setup}
\label{subsec:expdesc}

This study investigated the microstructural evolution during hot forging using a deformation dilatometer, specifically the TA Instruments' DIL805A/D/T model. This apparatus provides precise control over temperature and deformation, making it ideal for experimental simulations of the forging process. It allows for detailed monitoring and controlling of process variables which are challenging in actual forging processes. \\

In the pursuit of this study, the material chosen for the experiments was 18CrNiMo7-6 case-hardening steel, commonly used in gear components. This steel was selected due to its availability of a complete material model from a previous study performed by \citet{bambach2021soft}, which simplifies the analysis of the austenite grain size evolution. The experimental procedure involved cylindrical samples, measuring 5\,mm in diameter and 10\,mm in length, which were subjected to compression testing. \\

Transitioning to the testing phase, the compression tests were performed at two temperatures, 1180\,$^{\circ}$C and 1200\,$^{\circ}$C, with a constant strain rate of \(\sfrac{10}{s}\). These tests comprised three stages, each targeting different strains (0.3, 0.3, and 0.4) to accumulate a total strain of 1.0. The samples were heated under an argon gas atmosphere at a rate of 10\,\(\sfrac{K}{s}\) and maintained at temperature for 90\,s prior to compression. To explore the effects of temperature variations and control inputs, the study simulated disturbances by deviating the initial temperature from the target 1200\,$^{\circ}$C and adjusting wait times between the deformation stages. \\

For the actual compression process, silicon nitride (Si3N4) punches were employed. To minimize heat transfer from the specimen to the tools, thin molybdenum discs, measuring 0.1\,mm in thickness, were placed between them. Following each compression stage, rapid quenching of the samples was performed using argon gas at a cooling rate of 100\,\(\sfrac{K}{s}\). \\

For microstructural analysis, the specimens were bisected along their cylindrical axis and prepared by abrading on SiC papers of varying grit sizes and polishing with diamond suspensions and a silica solution. Special etching was applied to accentuate the prior-austenite grain boundaries. Light optical microscopy, using a Keyence Germany GmbH digital microscope and ATLAS image processing software (TESCAN software), was employed to measure the grain size. The target grain size for analysis was specified at 35\,\(\mu m\), with the aim to determine this boundary line within measurement windows of 490 \(\times\) 700\,\(\mu m\). \\

Lastly, it is worth mentioning that there are limited options to interface the machine and implement own control algorithms. For this reason, the validation experiments simulate disturbances via deviations of the initial temperature from the target value of 1200\,\textdegree{}C, and implement property control by computing adapted wait times between the three forming steps. 

\subsection{Dataset Generation}

A dataset for trainung was created using 500 forward simulations with the FE simulation described in Subsection \ref{sec:hotworkingprocess}. In addition, this simulation was divided into individual strokes, so that the dataset consists of a total of 4,000 input-output pairs. These data were split in a ratio of 80:10:10 between training, validation and test datasets. Finally, the simulations were run on an Intel Core i9-10900K. \\

Table \ref{tab:limits} presents the limits of the parameter range that define the forging strategy used for all the simulations.

\begin{table}[H]
\caption{Limit values used as input for the simulation, between which random values were sampled.}
\centering
\begin{tabular}{lcccc}
\hline
         & Temperature {[}\textdegree{}C{]} & \begin{tabular}[c]{@{}c@{}}Transport \\ time {[}s{]}\end{tabular} & \begin{tabular}[c]{@{}c@{}}Wait \\ time {[}s{]}\end{tabular} & \begin{tabular}[c]{@{}c@{}}Upsetting \\ time {[}s{]}\end{tabular} \\ \hline
Oven     & 1100--1300                       & 0--30                                                             & --                                                           & --                                                                \\
Stroke 1 & --                               & --                                                                & 1--60                                                        & 0.05--0.15                                                        \\
Stroke 2 & --                               & --                                                                & 1--60                                                        & 0.05--0.15                                                        \\
Stroke 3 & --                               & --                                                                & 1--60                                                        & 0.05--0.15                                                        \\ \hline
\end{tabular}
\label{tab:limits}
\end{table}

Table \ref{tab:scalesdeep} shows the limit values and ranges of all six 2D arrays building the final dataset, i.e., temperature, grain size, recrystallization, deformations in x and y direction and equivalent plastic strain. Temperature expresses the temperature to which the oven from Figure \ref{fig:deepforgefemschema} is preheated. Consequently, a workpiece is heated up to this temperature.

\begin{table}[H]
\centering
\caption{Limit values and ranges for the 2D arrays, namely temperature, grain size, recrystallization, deformations in x and y direction denoted as \textit{$U_x$} and \textit{$U_y$} and equivalent plastic strain denoted as \textit{EQPlast}.}
\begin{tabular}{@{}lccc@{}}
\toprule
       & Temperature [°C] & $U_x$ {[}mm{]}             & $U_y$ {[}mm{]}       \\ \midrule
Limits & 938 -- 1450     & 0 -- 2.41              & -5 -- 0         \\
Range  & 512              & 2.41                   & 5               \\ \midrule
       & EQPlast {[}-{]}  & Recrystallization {[}-{]} & Grain size [$\mu$m] \\ \midrule
Limits & 0 -- 1.13        & 0 -- 1                    & 17.5 - 70           \\
Range  & 1.13             & 1                         & 52.5                \\ \bottomrule
\end{tabular}
\label{tab:scalesdeep}
\end{table}

Moreover, all input parameters except for the temperature and grain size are at the beginning of the forging process equal to 0. Temperature is randomly sampled within the limits shown in Table \ref{tab:scalesdeep} and grain size is set to 70 $\mu m$. \\

\begin{figure}[H]
\centering
\SetFigLayout{1}{2}
  \subfigure[Example of an input.]{\includegraphics[width=0.48\textwidth]{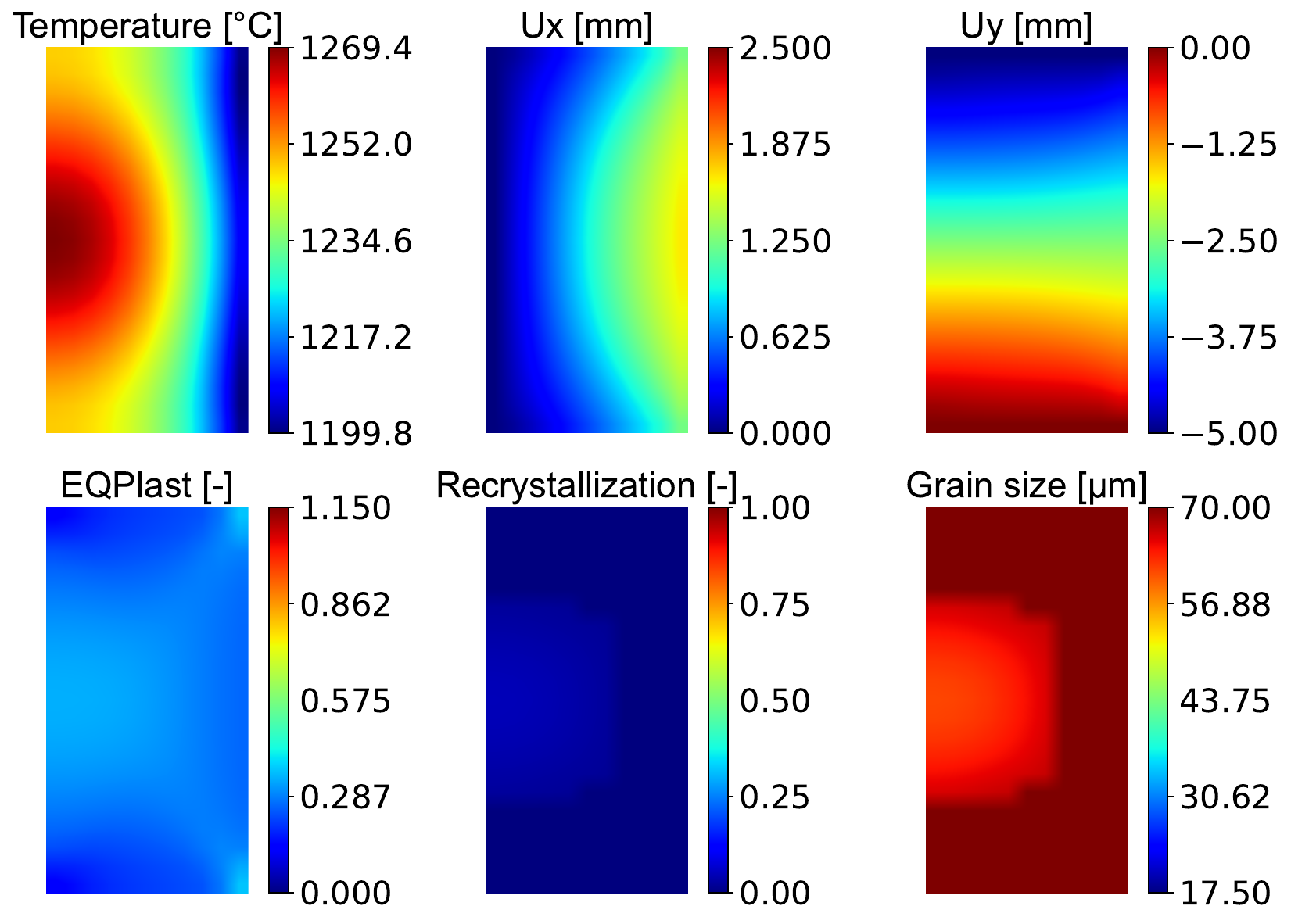}
    \label{fig:inputsa}}
  \hfill
  \subfigure[Coresponding output after a stroke]{\includegraphics[width=0.48\textwidth]{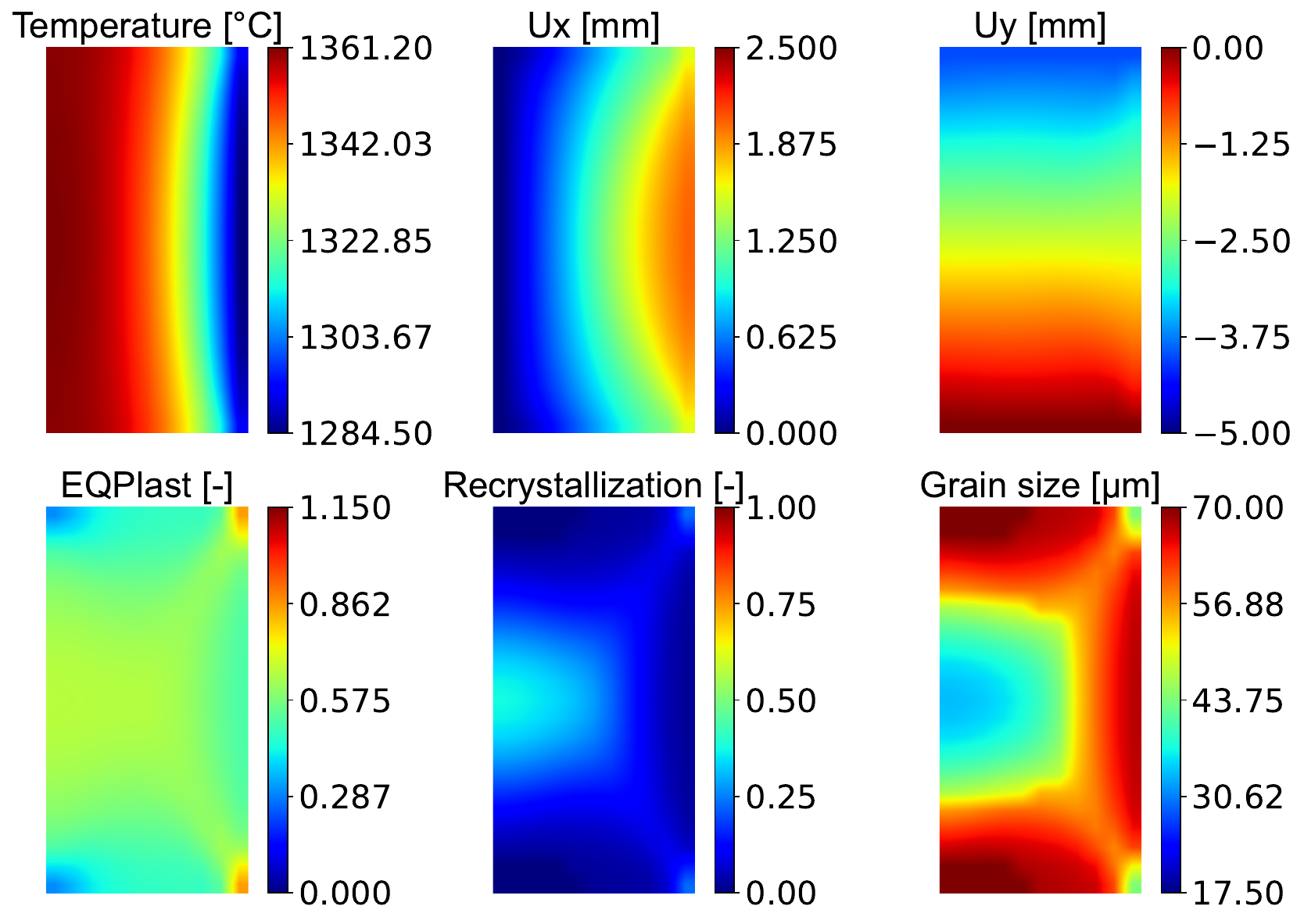}
    \label{fig:inputsb}}
  \caption{Example of an input-output pair, where the applied stroke is characterized with the wait time of 7.6\,s and upsetting time of 0.13\,s.}
  \label{fig:pairs}
\end{figure}

Finally, Figure \ref{fig:pairs} shows an example of the input-out pair of workpiece microstructure, i.e. before and after a stroke. The pair consist of twelve 2D arrays of temperature, deformation in x and y, equivalent plastic strain, recrystallization and grain size.

\subsection{Data Preprocessing}

The processing of the input data involves the following steps:

\begin{itemize}
    \item Normalization of all the 2D arrays representing the internal specimen state between 0 and 1.
    \item Normalization of all the 3D vectors representing the forging strategy between 0 and 1.
    \item Reshaping the 2D arrays into the shape of (batch size, 6, 21, 11), where the channels of the second dimension correspond to the microstructure properties.
    \item Extracting surface temperatures from the 2D arrays used as input of the machine learning architecture.
    \item Creating input-output pairs, where the input is the surface temperature of a workpiece and the output is its microstructure affected by the stroke parametrized with upsetting time together with wait and transport times.
\end{itemize}

\begin{figure}[H]
    \centering
    \includegraphics[width=0.5\textwidth]{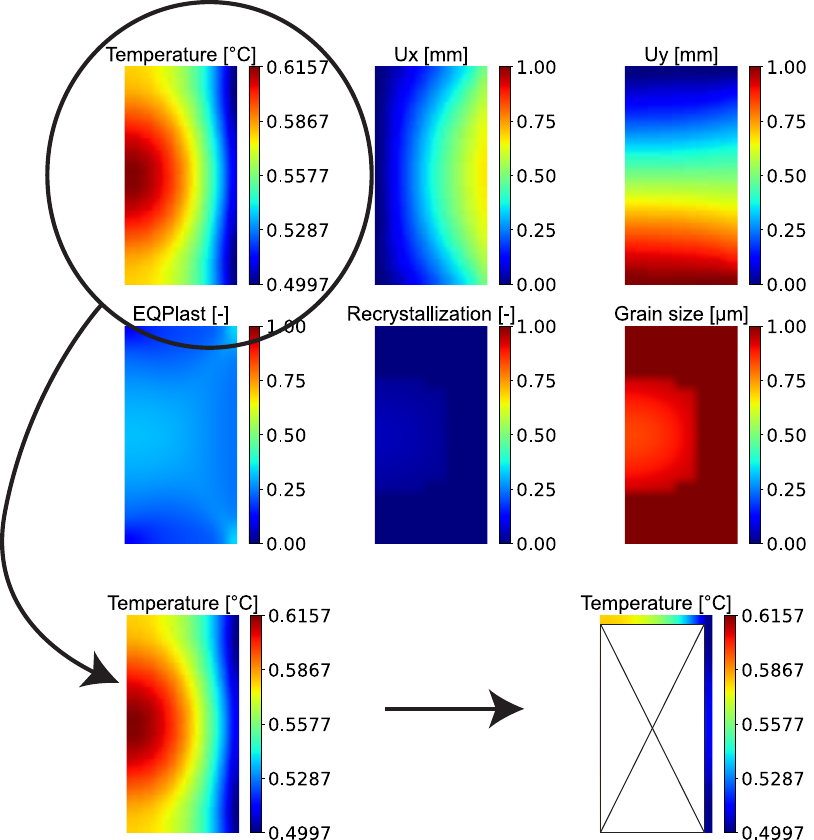}
    \caption{Visualisation of the input required for DeepForge, where only the normalized surface temperature that can be measured in the real process is used to predict the full microstructure of a workpiece consisting of six 2D arrays.}
    \label{fig:inputdeep}
\end{figure}

Figure \ref{fig:inputdeep} shows the extraction of the surface temperature from Figure \ref{fig:inputsa}, which serves as the input to the machine learning architecture and based on which are predicted all six 2D arrays of the microstructure after each stroke. The surface temperature, also called the temperature contour, was chosen to provide an input to the machine learning architecture that is measurable during the process itself.

\subsection{DeepForge Architecture}

\begin{figure}[H]
    \centering
    \includegraphics[width=1.0\textwidth]{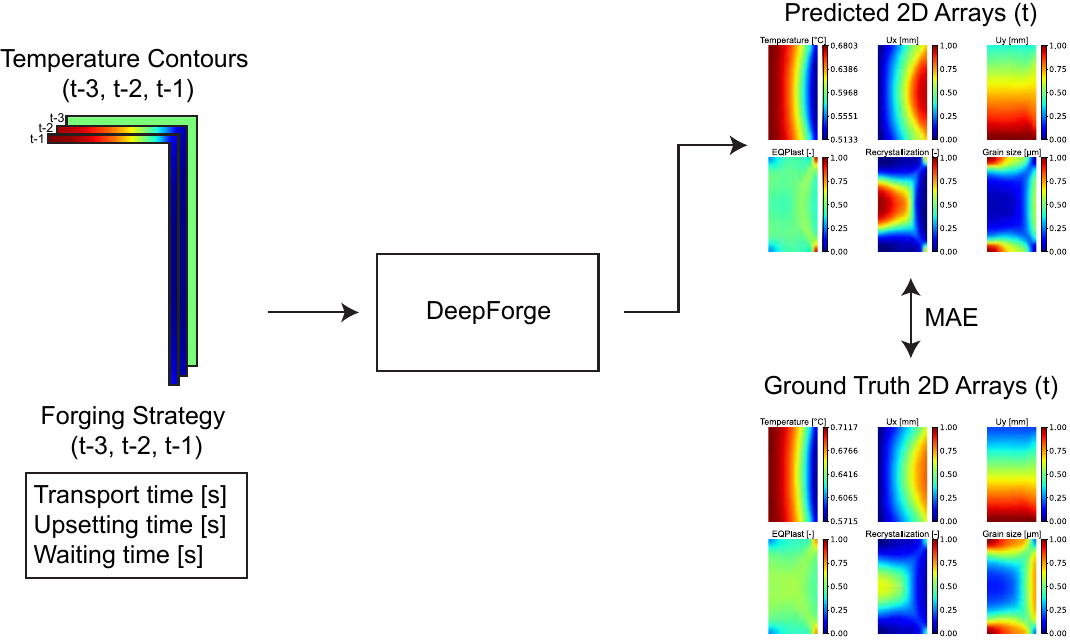}
    \caption{The general workflow of DeepForge in the train scenario involves taking surface temperatures, i.e. surface temperatures, from time moments $t-3$, $t-2$, $t-1$ along with a forging strategy as input. It is important to highlight that in instances where prior measurements are absent, such as when a part is freshly removed from an oven and undergoes deformation during the first forging stroke, the first two surface temperatures are assigned a default value of 0. This input information is then passed through DeepForge, which ultimately reconstructs 2D arrays expressing the microstructure of the part, incorporating the effects of the forging strategy applied. Finally, the reconstructed arrays are compared to the ground truth using the Mean Absolute Error (MAE) loss function. This loss is backpropagated and the entire network is trained over a number of epochs to optimise the performance of the model.}
    \label{fig:deepforgeworkflow}
\end{figure}

The general training workflow of the DeepForge is shown in Figure \ref{fig:deepforgeworkflow}.

The DeepForge machine learning framework processes surface temperatures and forging vectors from time moments $t-3$, $t-2$, $t-1$, and forecasts six two-dimensional arrays representing the microstructure of a workpiece in a hot forging process at time $t$. The accuracy of these predictions is assessed using a mean absolute error (MAE) loss function, which compares the predicted arrays with the ground truth data to measure the precision of the reconstruction. \\

The architecture of the DeepForge is shown in Table \ref{tab:decoderarchitecture_DP}.

\begin{table}[H]
\centering
\caption{The architecture of the DeepForge, incorporating surface temperatures of the last three forging strokes (bs, 32, 3) and the forging strategy as vector of these strokes (bs, 3$\times$3) where bs stands for batch size. Furthermore each operation is followed by a ReLU activation function and between the Linear layers is also a dropout with probability of 0.1. Finally, the kernel size of the Conv1D layers is always 3$\times$3 with padding of 1 and stride of 2$\times$2.}
\begin{tabular}{@{}ccc@{}}
\toprule
Operation & Input Type                                                   & Output Shape  \\ \midrule
Conv1D    & Contours                                                     & (bs, 8, 3)    \\
Conv1D    & Contours                                                     & (bs, 16, 3)   \\
Conv1D    & Contours                                                     & bs, 32, 3)    \\
GRU       & Contours                                                     & (bs, 32, 16)   \\
Flatten   & Contours                                                     & (bs, 512)     \\
Concat    & \begin{tabular}[c]{@{}c@{}}Contours\\ Strategy\end{tabular} & bs, 521)      \\
Linear    & Concat vec                                                   & (bs, 1024)    \\
Linear    & Concat vec                                                   & (bs, 512)     \\
Linear    & Concat vec                                                   & (bs, 1024)    \\
Linear    & Concat vec                                                   & (bs, 6$\times$21$\times$11) \\ \bottomrule
\end{tabular}
\label{tab:decoderarchitecture_DP}
\end{table}

The training parameters are written in Table \ref{tab:trainparam_DP}.

\begin{table}[H]
\centering
\caption{Training parameters of DeepForge.}
\begin{tabular}{@{}cccc@{}}
\toprule
Batch size & Learning rate & Epochs & Optimizer \\ \midrule
128         & 0.001        & 1,000  & Adam     \\ \bottomrule
\end{tabular}
\label{tab:trainparam_DP}
\end{table}

These training parameters were chosen based on a combination of empirical evidence and best practices within the field of machine learning. Selecting a batch size of 128 balances computational efficiency and generalization regarding the convergence speed and stability. Moreover, the learning rate of 0.001 is a common default for the Adam optimizer. Ultimately, the decision to use 1,000 epochs was based on the model consistently achieving convergence prior to reaching this number.

\subsection{Model Predictive Control}
\label{subsec:mpc}

Model predictive control (MPC) is a method of process control that has been widely used in various industrial applications due to its ability to handle multivariable control systems with constraints. The MPC framework operates by creating a model to predict future behavior and optimize the performance of the control system over a finite time horizon (see \citet{garcia1989model}). \\

\begin{figure}[H]
    \centering
    \includegraphics[width=1.0\textwidth]{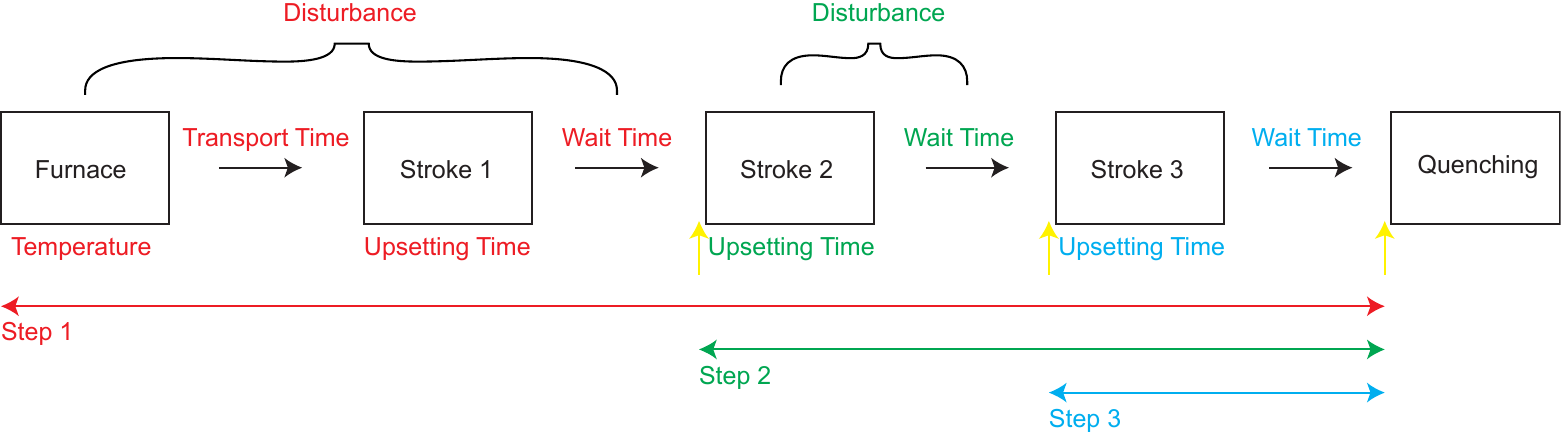}
    \caption{Schema of the MPC framework for control of three-stroke hot closed-die forging process (see Subsection \ref{sec:hotworkingprocess}). Thee yellow arrows visualise the moments when the DeepForge outputs the microstructure of the workpiece parametrized on the surface temperatures from three previous time moments.}
    \label{fig:mpc_schema}
\end{figure}

In the context of a forging process, MPC is effective due to its predictive capabilities, which allow for anticipation and compensation for disturbances that may adversely affect the process outcomes. Figure \ref{fig:mpc_schema} illustrates the framework of the model predictive control (MPC) system which optimizes a total of eight parameters, denoted by \( u_1, u_2, \ldots, u_8 \). This setup is designed to be responsive to any disturbances that may occur before the second stroke and between the beginning of the second and third stroke.

The MPC workflow progresses through three main steps, targeting the minimization of a predefined objective function, \( J(u) \), at each stage:

\begin{enumerate}
    \item All 8 parameters $u_1, u_2, \ldots, u_8$ (in red, green and blue) are optimized simultaneously to obtain the optimal value of the objective function \( J(u) \) after the third forging stroke. Upon reaching optimal solutions, the parameters for the initial forging stroke, indicated in red, are fixed and no longer used.
    \item The MPC workflow re-runs from the second stroke on. It allows for an update to the input state, which may be necessitated by disturbances in the process or availibility of more precise measurement data. The optimisation at this stage focuses on the remaining four parameters $u_5, u_6, u_7, u_8$ (depicted in green and blue) to refine the value of the objective function. Once optimal solutions are found, the parameter values determined in this step are fixed.
    \item The workflow runs the last time for the third stroke where the last set of parameters $u_7, u_8$ (shown in blue) are optimized and determined. This final optimisation step also permits updates to the input, ensuring that the parameters are fine-tuned based on the latest state of the system that might have changed because of additional disturbances.
\end{enumerate}

This means that there is an adaptive prediction horizon within the MPC framework, where the optimisation process dynamically adjusts its focus on the control parameters through the forging sequence. Initially, all three steps consisting of eight parameters are optimised, but as the process progresses, the horizon narrows and focuses on fewer parameters in response to the evolving state of the system. This adaptive approach aims at pre-optimising the parameters in the following steps. Furthermore, the control horizon is always one step, as the parameters of only one step are applied at one time before moving on to the next.\\

The objective function \( J(u) \) used for optimisation might, for instance, represent the degree of recrystallization after performing the three strokes or the evolution of grain size during the last wait time, i.e. after the third stroke. \\

For this publication, the objective function \( J(u) \), aimed at minimizing the occurrence of grain sizes exceeding 35 microns in a specified field $A$, is defined as follows:

\begin{equation}
J(u) = \sum_{i,j} f_{\text{thresh}}(A_{ij}, n) + k \times \sum_{i} WT_{i}
\end{equation}

where $WT_{i}$ is the $i$th wait time, k is the weight coefficient equal to 0.3 and \( A \) is matrix representing the 2D field of interest. Furthermore, $f_{thresh}$ is a threshold function, which is given as follows:

\begin{equation}
    f_{\text{thresh}}(x, n) = \begin{cases}
    1 & \text{if } x \leq n, \\
    0 & \text{if } x > n.
    \end{cases}
\end{equation}

In this function, $n$ is a threshold parameter, which is set to be 35 microns. \\

\begin{figure}[H]
    \centering
    \includegraphics[width=0.5\textwidth]{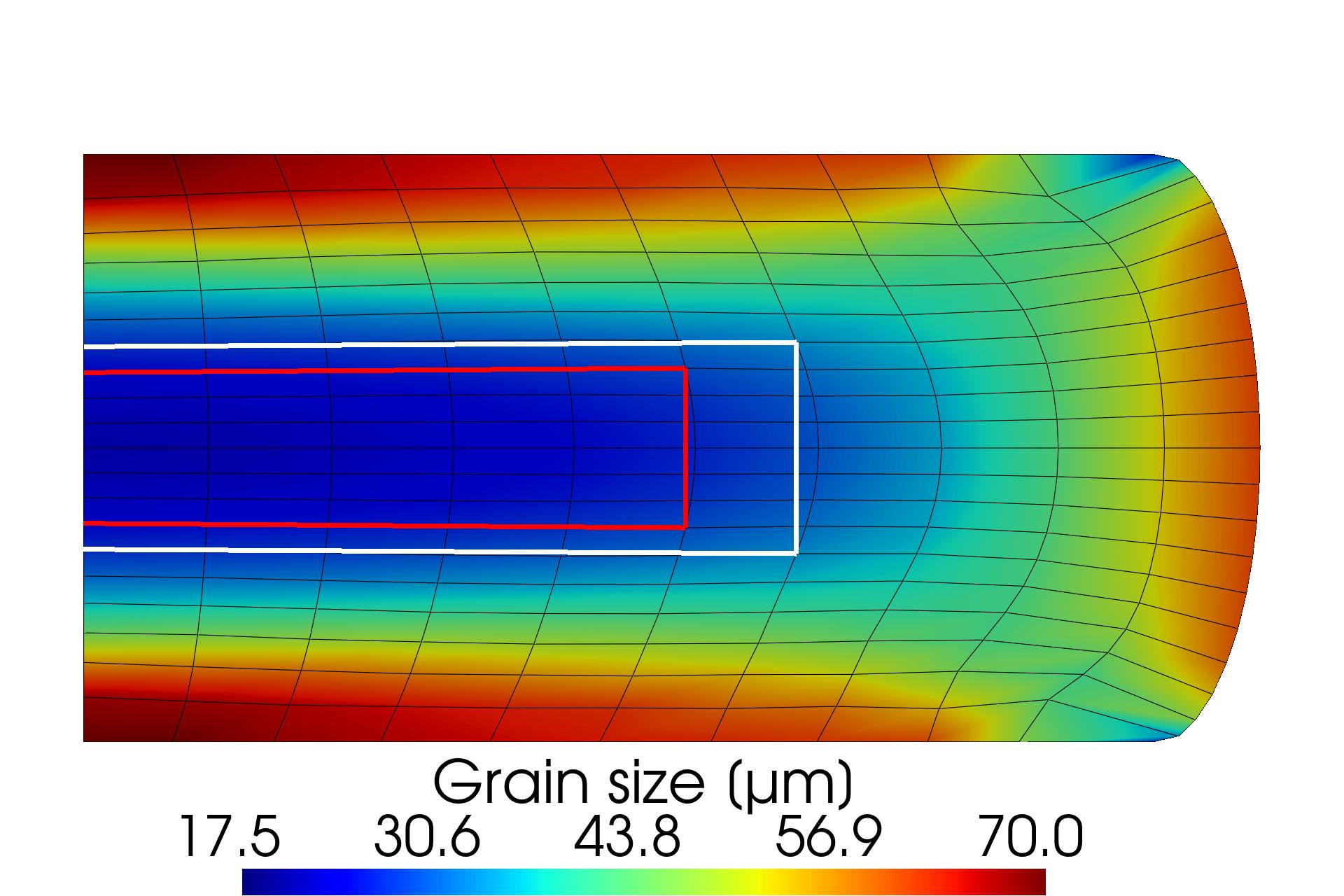}
    \caption{Objective function for maintaining grain size below 35 microns, where the inner area of the red block represents the 2D field of interest and the white line marks the safety margin around the field of interest where the grain size shall be under 35 microns.}
    \label{fig:objective function}
\end{figure}

Finally, the objective function \( J(u) \) is visualized in Figure \ref{fig:objective function}.

The forging strategy is optimized using stochastic optimisation algorithm, specifically speaking dual annealing to minimize the objective function (see \citet{xiang2013generalized}). \\

The hyperparameters used within the dual annealing algorithm are summarized in Table \ref{tab:gaparams}.

\begin{table}[H]
\centering
\caption{Hyperparameters of the dual annealing algorithm used for the optimisation of the forging strategy  according to \cite{storn1997differential}.}
\begin{tabular}{@{}cccccc@{}}
\toprule
Max iterations & Max evaluations & Initial Temp. & Visit & Accept \\ \midrule
   300             & $1\times10^7$ &  10,000            &   2.7            &  -10   \\ \bottomrule
\end{tabular}
\label{tab:gaparams}
\end{table}

\section{Results}
\label{sec:results}

\subsection{Qualitative Results}

\begin{figure}[H]
\centering
\SetFigLayout{2}{2}
  \subfigure[Prediction for case number 1.]{\includegraphics[width=0.48\textwidth]{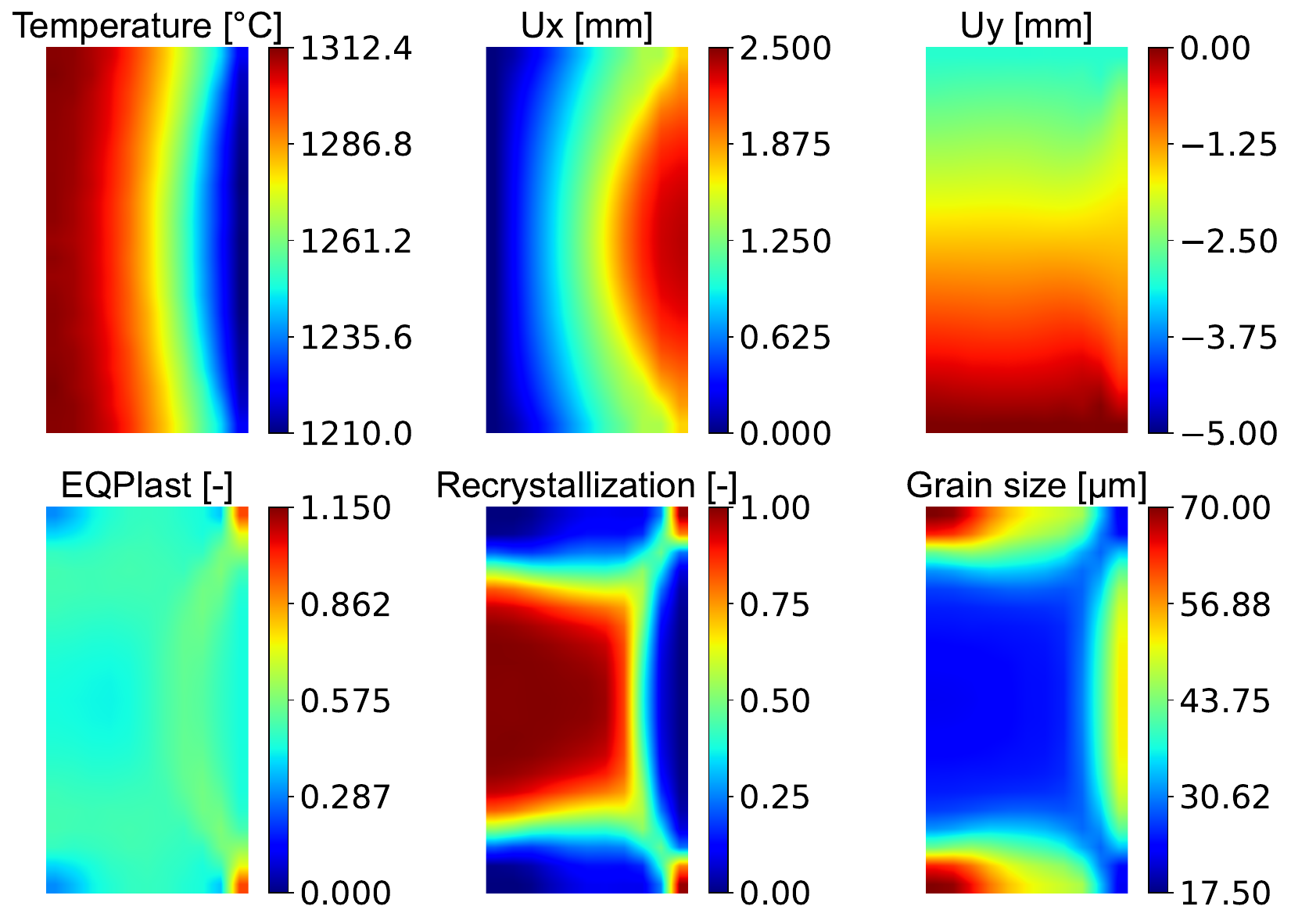}
  \label{fig:22a}}
  \hfill
  \subfigure[Prediction for case number 2.]{\includegraphics[width=0.48\textwidth]{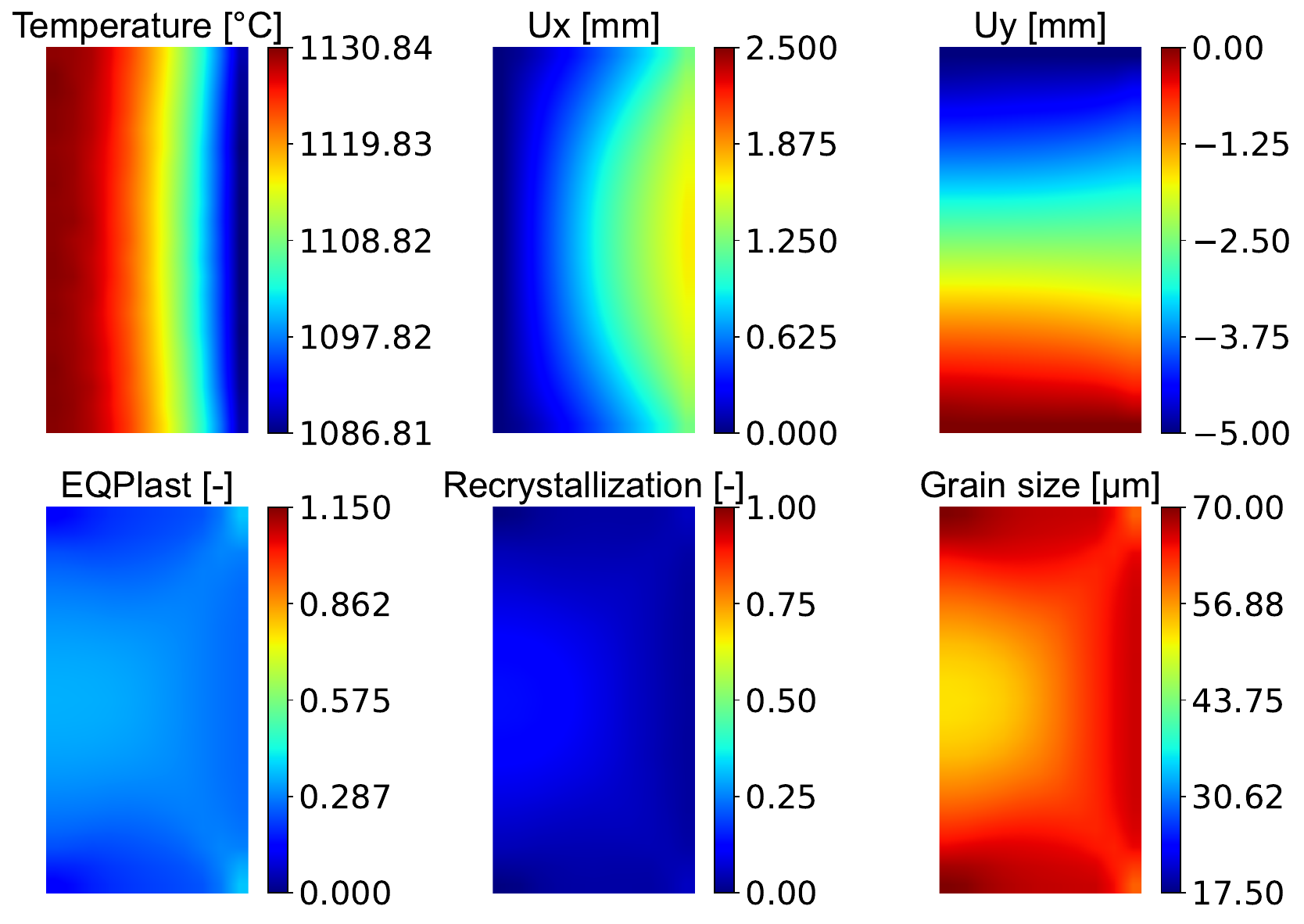}
  \label{fig:22b}}
    \hfill
  \subfigure[Comparison for case number 1.]{\includegraphics[width=0.48\textwidth]{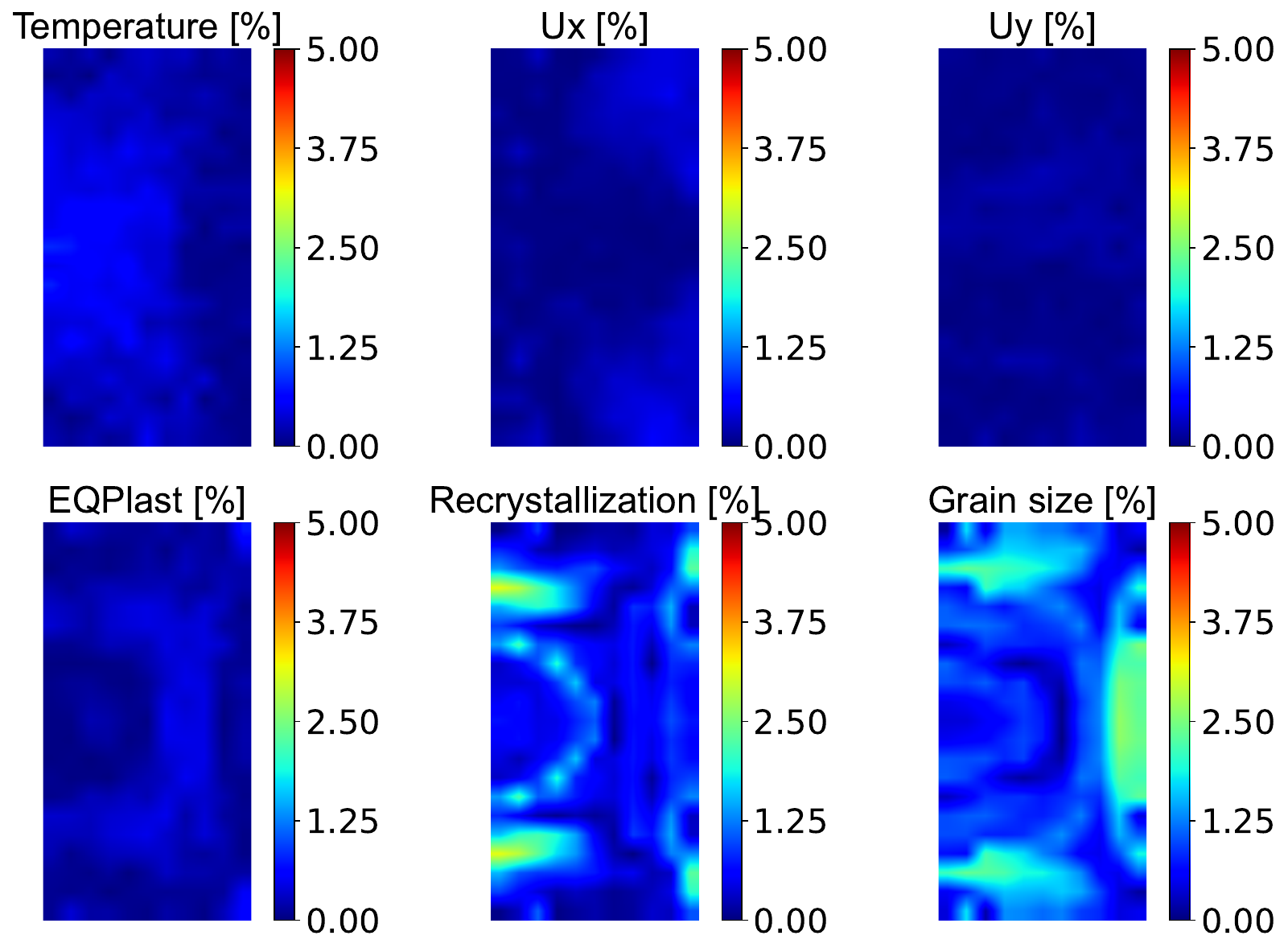}
  \label{fig:22c}}
  \hfill
  \subfigure[Comparison for case number 2.]{\includegraphics[width=0.48\textwidth]{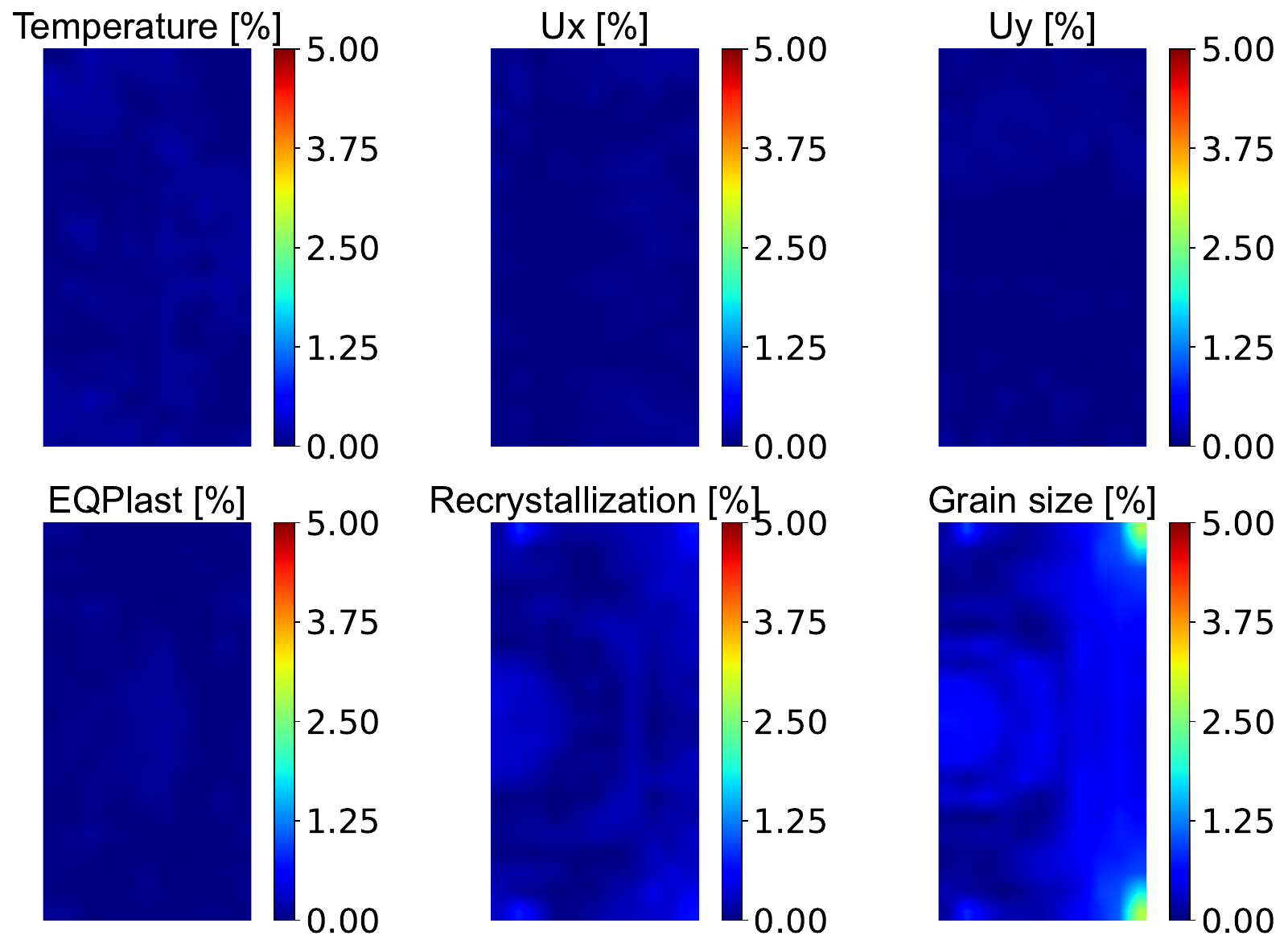}
  \label{fig:22d}}
    \hfill
  \caption{Visualisation of two cases where DeepForge predicted the 2D microstructure of a workpiece after a stroke based on the surface temperatures of the workpiece before the stroke. The case number 1 shown in Figure \ref{fig:22a} visualizes the prediction after third stroke stroke where the oven temperature was 1263\,\textcelsius.\:Furthermore, the transport time was 17\,s, the wait times were 11.7, 12.1, 16\,s and the upsetting times were 0.09, 0.1, 0.12\,s. As far as the case number 2 is concerned (see Figure \ref{fig:22b}) visualizes the results of microstructure prediction after the third stroke where the oven temperature was 1181\,\textcelsius\:the transport time was 0.05\,s, the wait time was 40.7\,s and the upsetting time was 0.09\,s. Finally, as shown in Figures \ref{fig:22c} \ref{fig:22d} a comparison of these predictions with the FE-based simulations is also provided in terms of percentage errors.}
  \label{fig:res2}
\end{figure}

The qualitative results for two cases in terms of DeepForge prediction of all six arrays representing the microstructure of a workpiece together with their comparison with the FE-based simulation are shown in Figure \ref{fig:res2}. These predictions were made based solely on the workpiece surface temperatures of the previous three strokes, as shown in Figure \ref{fig:deepforgeworkflow}.

\subsection{Quantitative Results}

Table \ref{tab:quantres} presents the quantitative results for each of the three strokes individually and for the overall performance, detailing the mean absolute error and standard deviations.

\begin{table}[H]
\centering
\caption{Quantitative results for the overall performance, detailing the mean absolute errors and standard deviations.}
\begin{tabular}{@{}cccccc@{}}
\toprule
Temperature {[}\(^{\circ}\)C{]}& $U_{x}$ {[}mm{]} & $U_{y}$ {[}mm{]} & EQPlast {[}--{]} & Recrystallization {[}--{]} & Grain size   {[}$\mathrm{\mu m}${]}  \\ \midrule
     2.1$\pm$3.2 & 0.003$\pm$0.002       &  0.006$\pm$0.005   &  0.002$\pm$0.002           & 0.005$\pm$0.006    &  0.4$\pm$0.3    \\ \bottomrule
\end{tabular}
\label{tab:quantres}
\end{table}

Furthermore, the same table with percentage errors regarding the ranges of the data given in Table \ref{tab:scalesdeep} is shown in Table \ref{tab:quantrespercnt}.

\begin{table}[H]
\centering
\caption{Quantitative results for the overall performance, detailing the mean absolute errors and standard deviations.}
\begin{tabular}{@{}cccccc@{}}
\toprule
Temperature [\%]& $U_{x}$ [\%] & $U_{y}$ [\%] & EQPlast [\%] & Recrystallization [\%] & Grain size  [\%]  \\ \midrule
     0.4$\pm$0.6 &   0.1$\pm$0.09    &  0.1$\pm$0.09   &      0.2$\pm$0.2       & 0.5$\pm$0.7   &  0.8$\pm$0.7    \\ \bottomrule
\end{tabular}
\label{tab:quantrespercnt}
\end{table}

\subsection{Model Predictive Control}

\begin{figure}[H]
    \centering
    \includegraphics[width=0.9\textwidth]{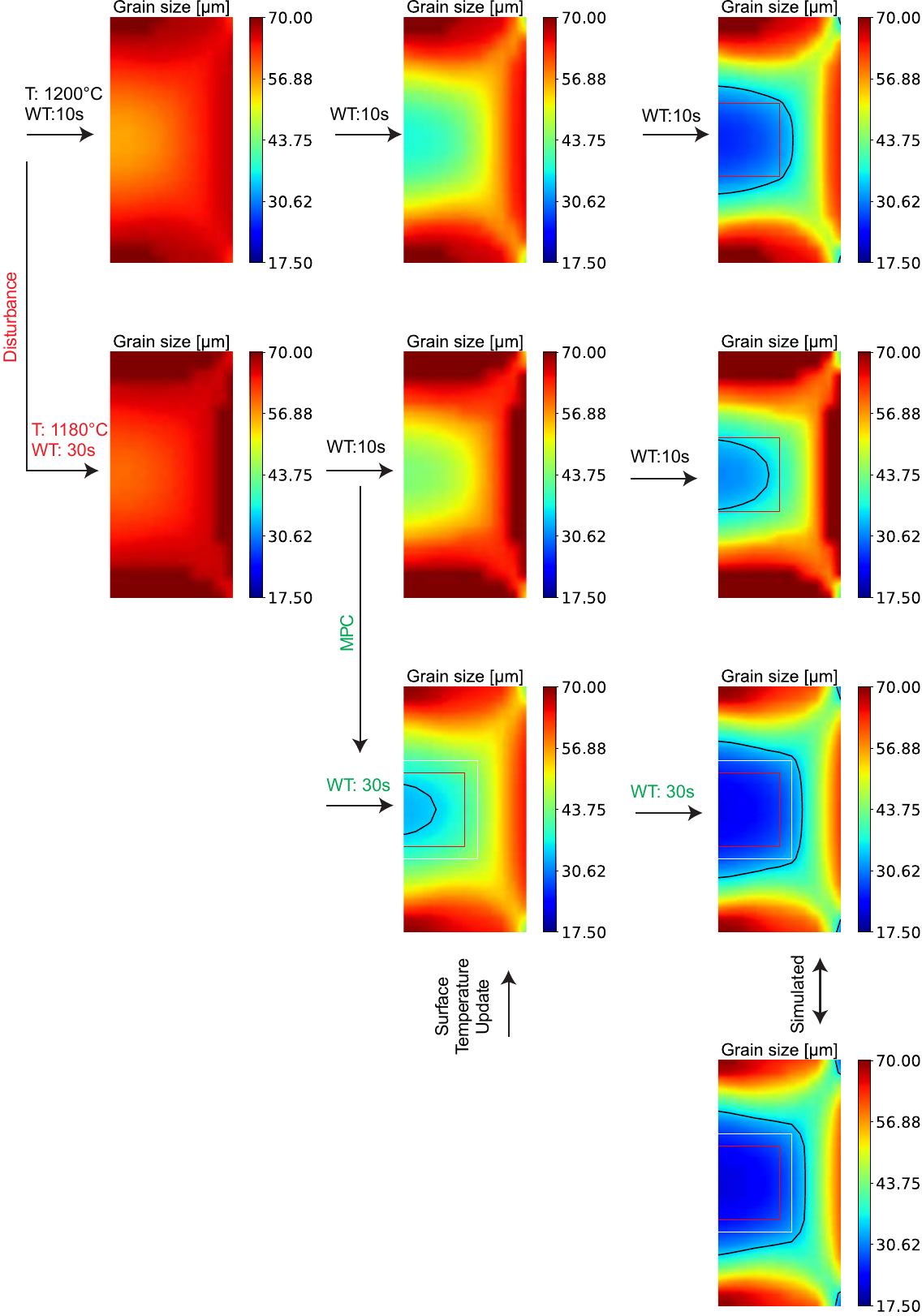}
    \caption{Visualisation of the MPC framework in the control of a forging process where disturbances occur in the initial temperature at which the furnace is heated. The aim is to achieve the desired grain size within a field with a safety margin represented by the white line. Please not that the presented results were rescaled to the original data ranges.}
    \label{fig:mpcres}
\end{figure}

Figure \ref{fig:mpcres} presents an MPC framework, adeptly achieving a desired grainsize in a workpiece undergoing forging, even in the face of disturbances such as variations in the initial temperature of the oven. Furthermore, the adjusted variable within this framework is the wait time, which is rounded to the nearest multiple of 5 to ensure a finite but sufficiently fine set of possible wait times.

The first scenario, as outlined in the first row, was planned using offline optimisation to achieve a grain size smaller than 35 microns in the field of interest. This was to be accomplished with a forging process involving an oven temperature of 1200\,\textdegree{}C and a 10-second wait time between successive strokes. However, as visualised in the second row, external factors caused the oven temperature to drop to 1180\,\textdegree{}C and extended the initial wait time to 30 seconds. This change meant the process no longer met the original goal of maintaining the grain size below 35 microns in the field of interest after the forging process. \\

In response, the MPC framework (detailed in Subsection \ref{subsec:mpc}) adjusted the strategy by increasing the subsequent wait times to 30 seconds (see the third row of Figure \ref{fig:mpcres}). This was done to not only achieve the desired grain size in the field of interest but also to add a safety margin, ensuring the grain size remains at 35 microns, while keeping the wait times to be low (see objective function in Subsection \ref{subsec:mpc}). \\

Then, in a real world scenario, there would be the possibility of updating the real surface temperature of the part, i.e. from an IR camera measurement. Here the surface temperature was predicted using FE simulation to model such an update. \\ 

The final result is visible in the last Figure of the third row. Below this resulting grain size field is the result from FE-based simulation (see Subsection \ref{sec:hotworkingprocess}) following the same forging strategy.

\subsection{Experimental Verification}

\begin{figure}[H]
\centering
\SetFigLayout{3}{3}
\subfigure[First Set: FE-based Simulation.]{\includegraphics[width=0.3\textwidth]{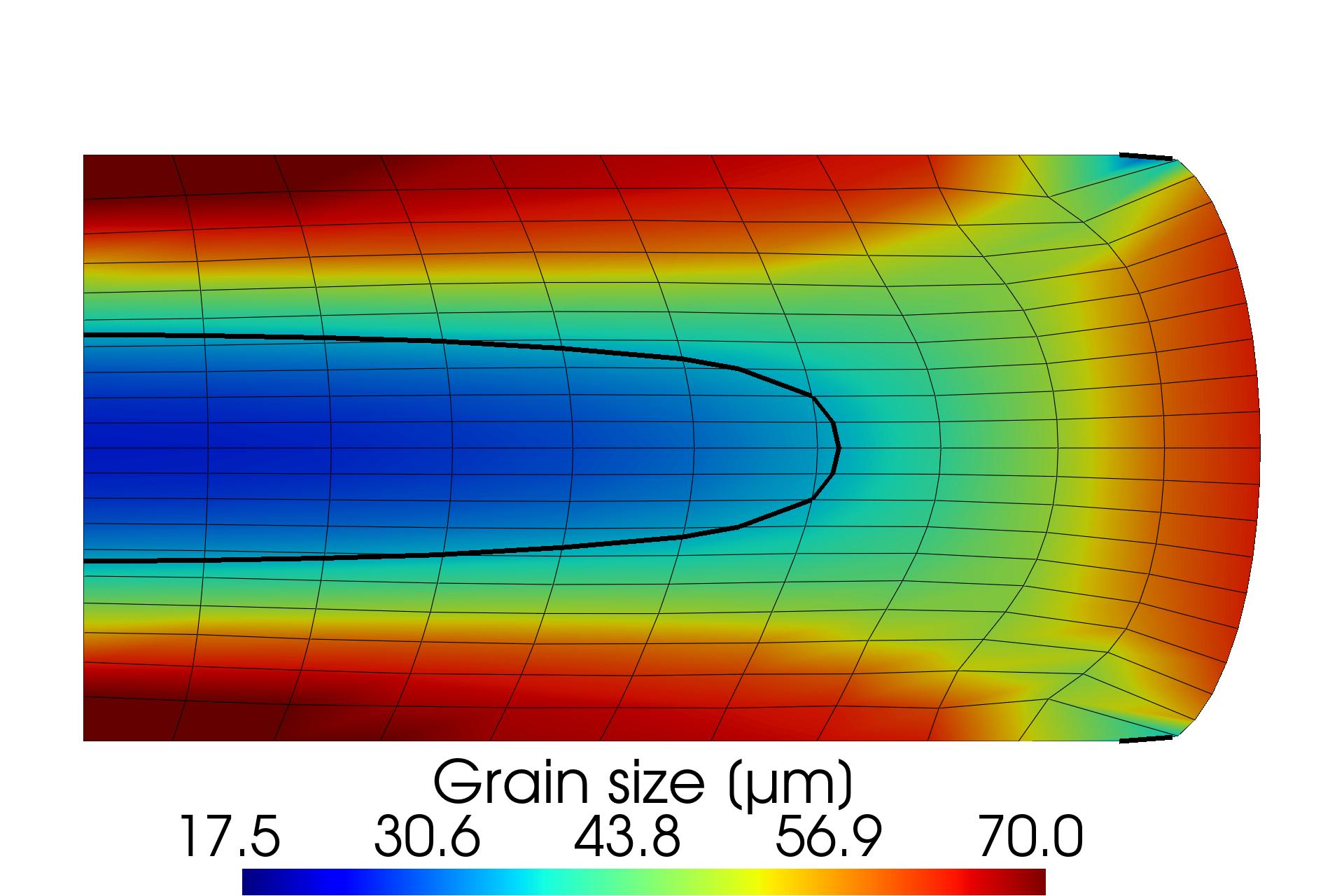}
\label{fig:23a}}
\hfill
\subfigure[First Set: DeepForge Prediction.]{\includegraphics[width=0.3\textwidth]{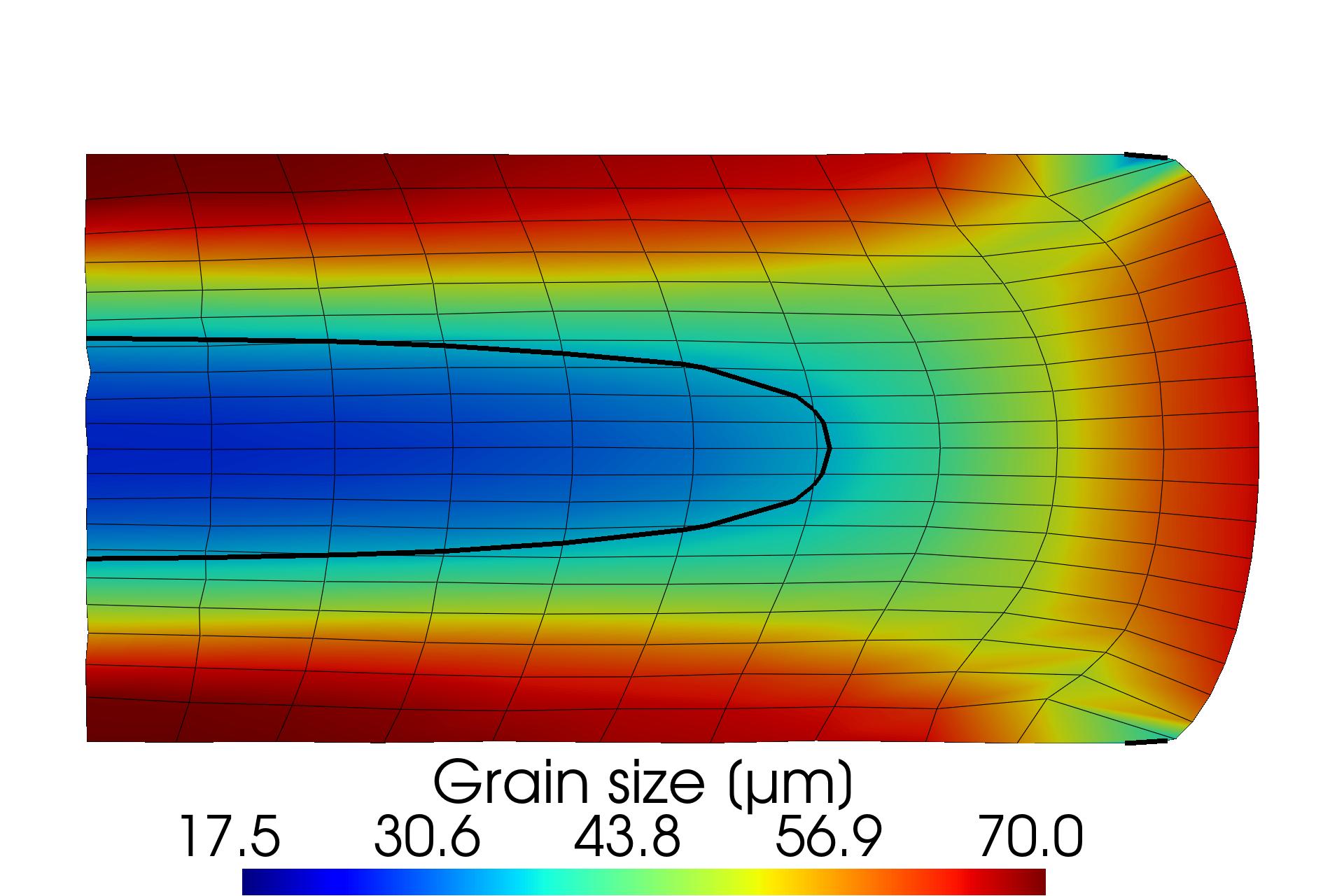}
\label{fig:23b}}
\hfill
\subfigure[First Set: Experimental Data.]{\raisebox{4mm}{\includegraphics[width=0.3\textwidth]{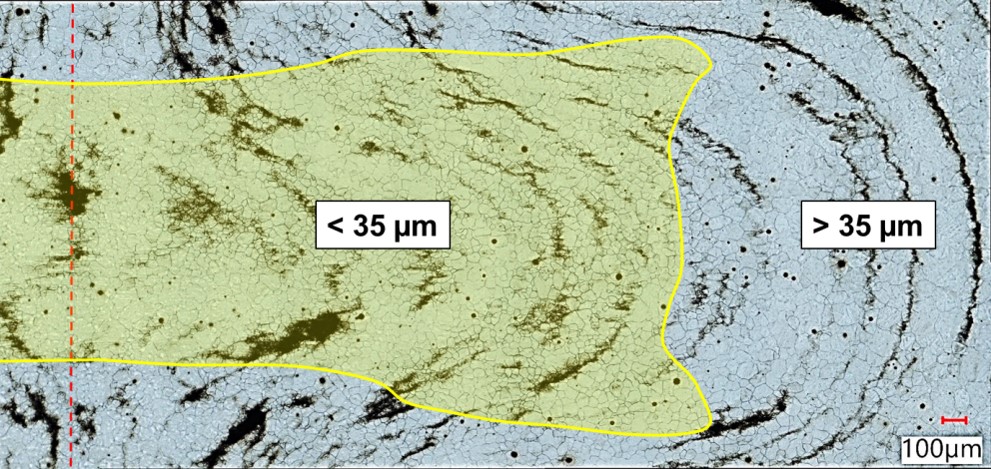}}
\label{fig:23c}}
\hfill
\subfigure[Second Set: FE-based Simulation.]{\includegraphics[width=0.3\textwidth]{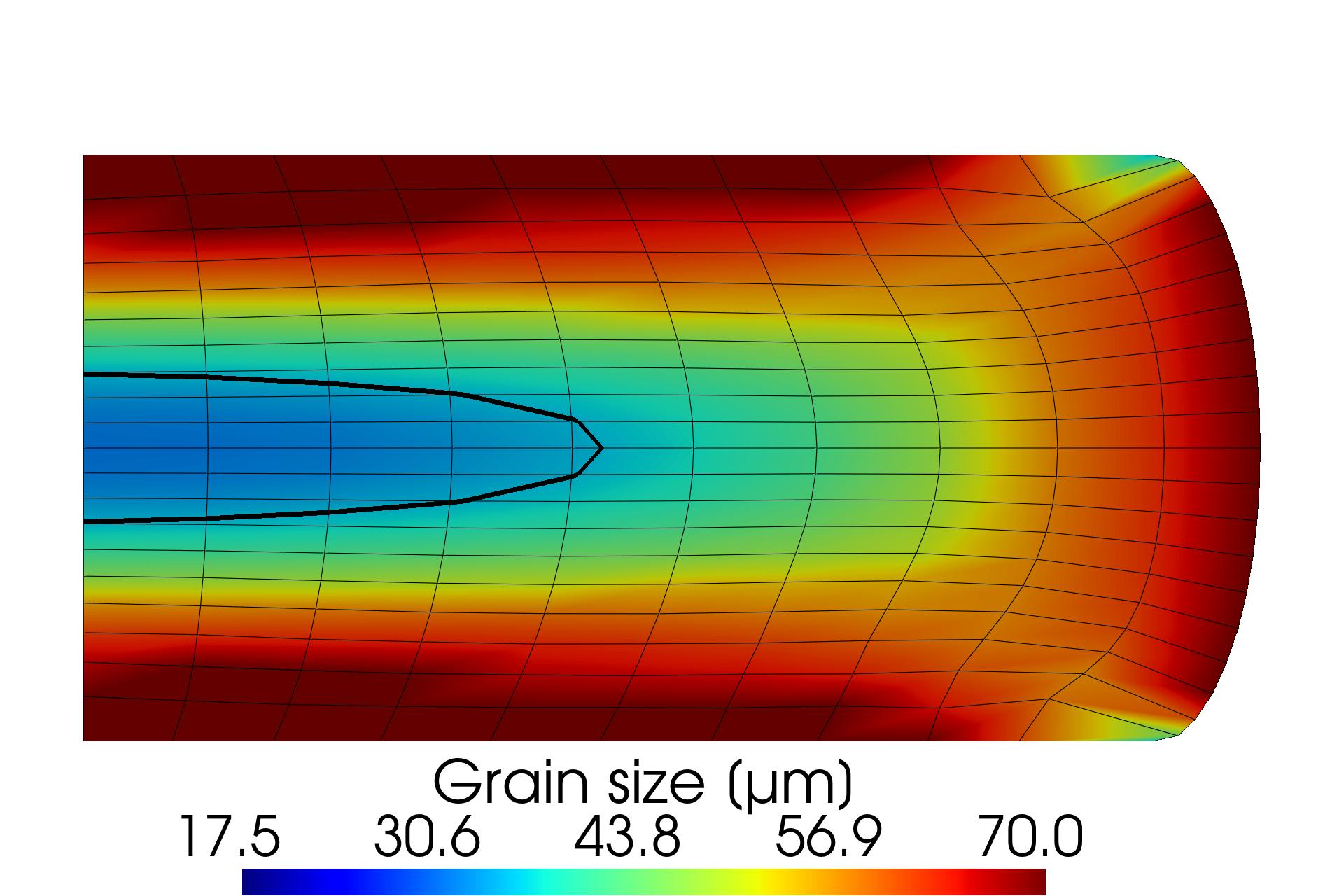}
\label{fig:23d}}
\hfill
\subfigure[Second Set: DeepForge Prediction.]{\includegraphics[width=0.3\textwidth]{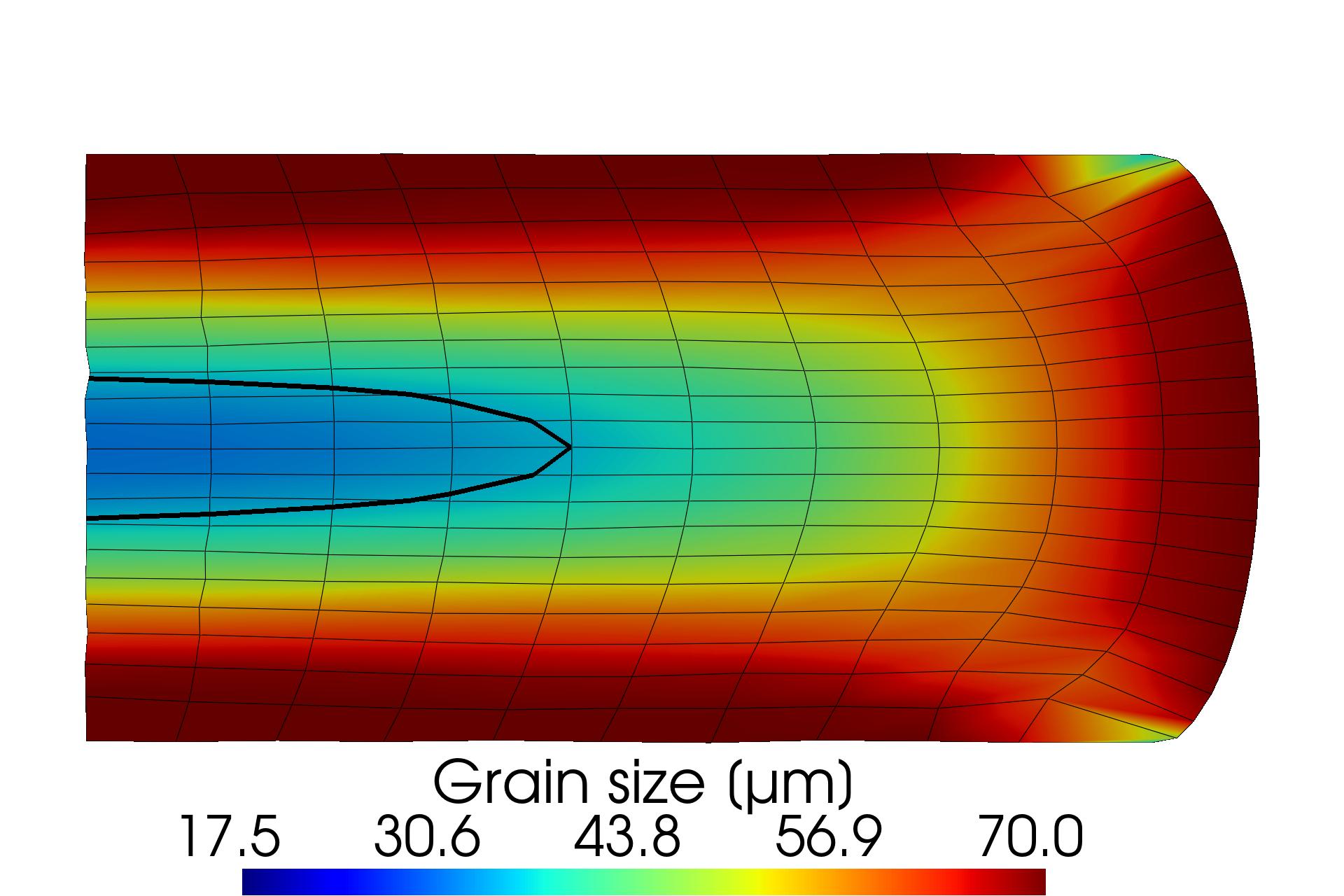}
\label{fig:23e}}
\hfill
\subfigure[Second Set: Experimental Data.]{\raisebox{4mm}{\includegraphics[width=0.3\textwidth]{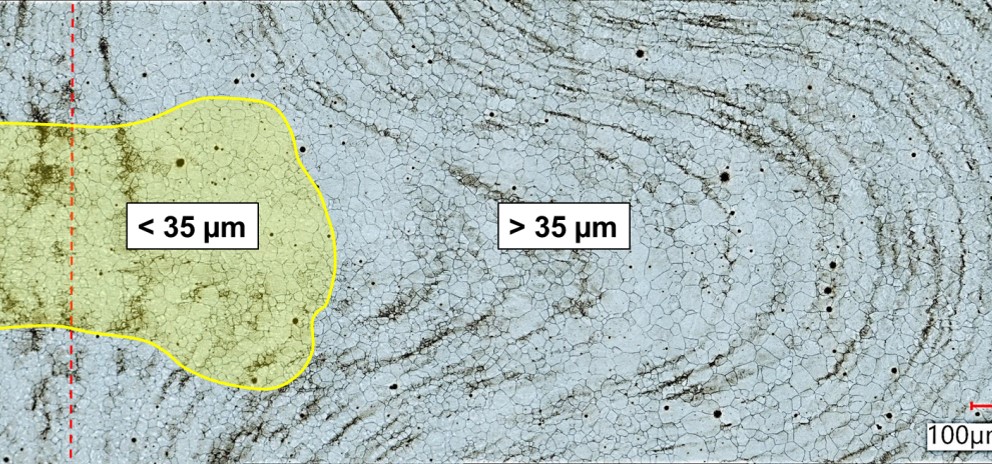}}
\label{fig:23f}}
\hfill
\subfigure[Third Set: FE-based Simulation.]{\includegraphics[width=0.3\textwidth]{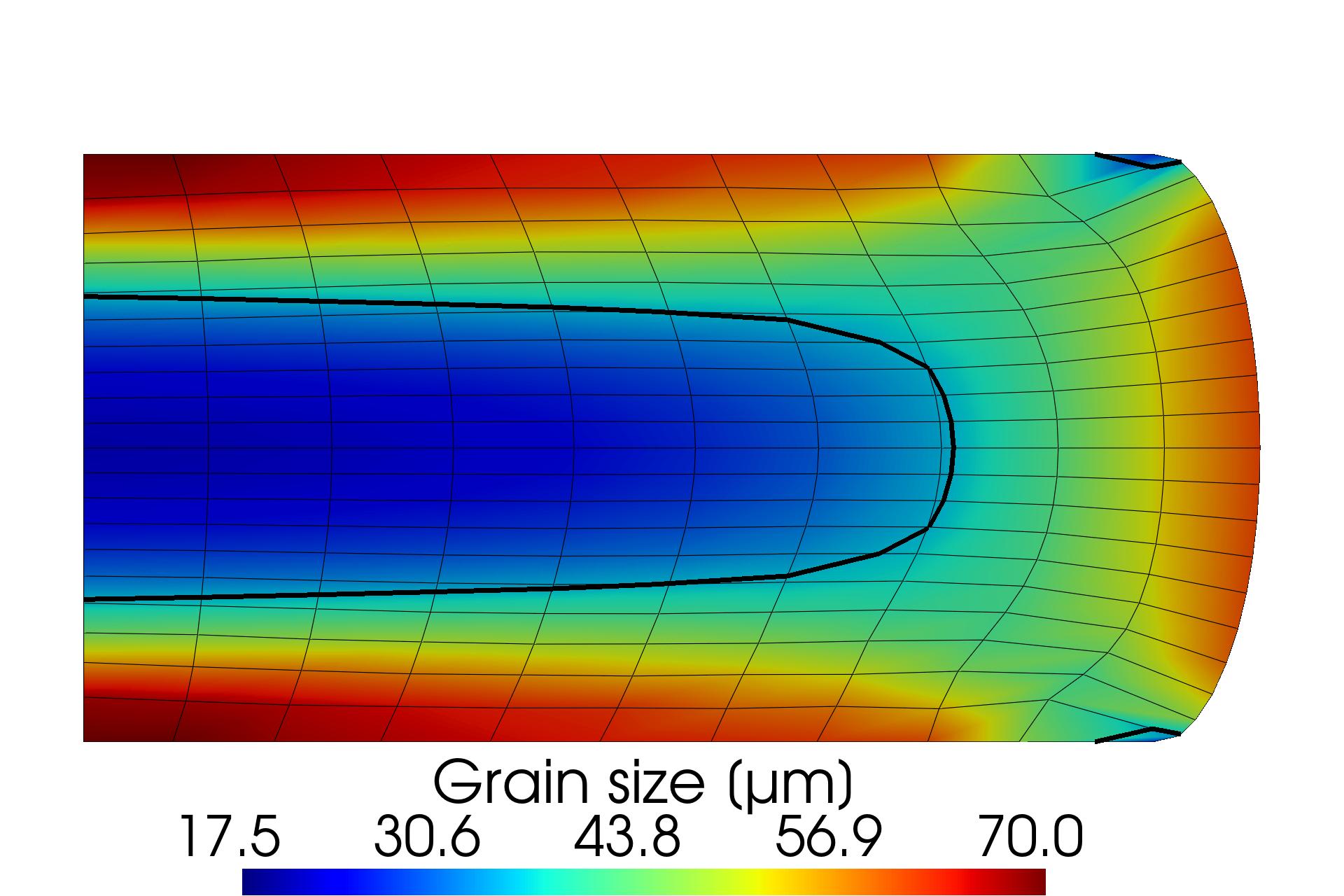}
\label{fig:23g}}
\hfill
\subfigure[Third Set: DeepForge Prediction.]{\includegraphics[width=0.3\textwidth]{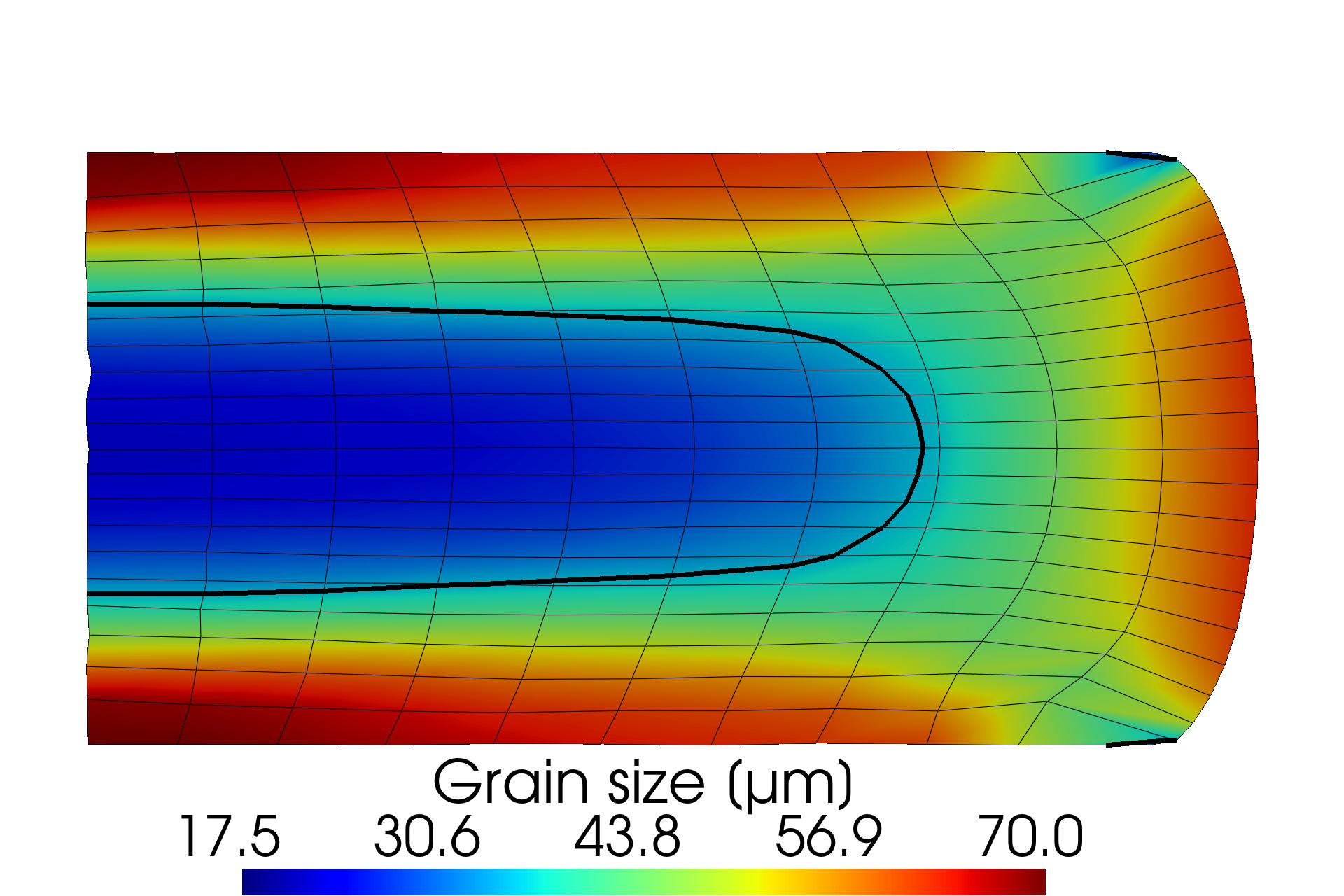}
\label{fig:23h}}
\hfill
\subfigure[Third Set: Experimental Data.]{\raisebox{4mm}{\includegraphics[width=0.3\textwidth]{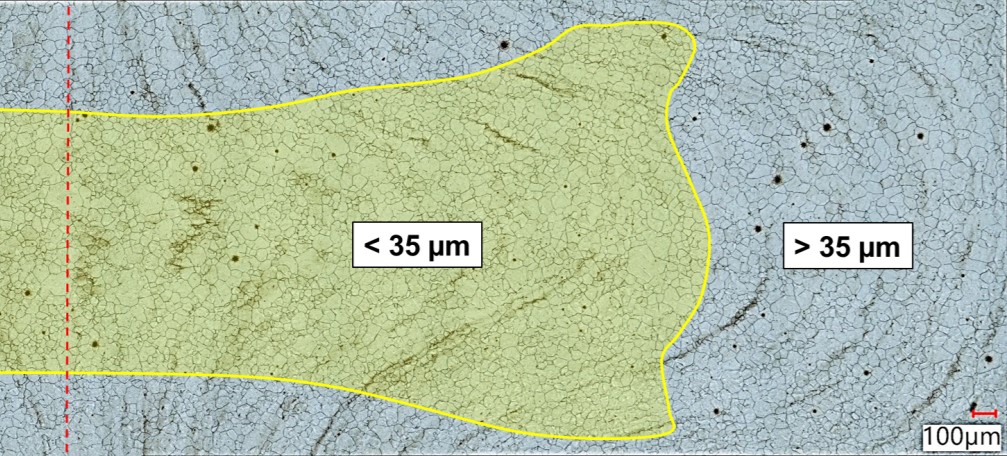}}
\label{fig:23i}}
\hfill
\caption{Grain size results for three forging strategies of FE-based simulation, DeepForge predictions, and experimental data. The top row shows results for a forging strategy with an oven temperature of 1200\,\textdegree{}C, no transport time and a wait time of 10 seconds. The middle row presents results for an identical strategy, but with an oven temperature of 1180\,\textdegree{}C. Finally, the last row has oven temperature of 1180\,\textdegree{}C and wait times of 30 seconds. The transport time is also equal to 0.}
\label{fig:res3}
\end{figure}

The initial as well as disturbed and optimized forging strategies from the MPC framework shown in Figure \ref{fig:mpcres} were experimentally verified using the procedure described in Subsection \ref{subsec:expdesc}. The results are summarized in Figure \ref{fig:res3}.

\section{Discussion}
\label{sec:discussion}


The DeepForge architecture, which uses a combination of 1D Convolutional Neural Networks and Gated Recurrent Units, is a strategic choice for use as a soft sensor in applications where full field output of a microstructure must be predicted from partial information, such as surface temperatures in forging processes.

The use of 1D CNNs is beneficial in this context where only partial data, i.e surface temperatures, is available, 1D CNNs can analyse this limited information to identify underlying patterns. This capability is essential to accurately infer the full field output from incomplete inputs. In addition, due to their convolutional nature, 1D CNNs are adept at handling data with spatial hierarchies, making them a good choice for surface temperature data where is naturally a spatial dependency.

Moreover, integrating GRUs into the architecture adds the ability to capture temporal dependencies and dynamics. This is particularly relevant in processes such as forging, where the state of the material and the outcome of the process are determined not only by the current conditions, but also by their evolution over time. In the case of DeepForge architecture, the input consist of the last three surface temperatures of a workpiece, based on which the microstructure prediction is done. It should be noted that if there are no previous measurements, for example if the part has just been taken out of an oven and is being deformed in the first forging stroke, the first two surface temperatures are simply set to 0.

Furthermore, the use of DeepForge for microstructure control in metal forming offers distinct advantages over traditional computational methods such as the Level Set Method (see \citet{osher1988fronts}). Firstly, it is rather intricate to find an accurate speed function $F$, because in order to follow the evolving grain size contour as shown in Figures \ref{fig:23g}, \ref{fig:23h} and \ref{fig:23i}, it must depend on the position on the contour \textbf{x} as well as time \textit{t} and oven temperature $T_{oven}$. In contrast, the definition of the speed function is not necessary for the development of DeepForge, which can effectively handle both initial input variations and perturbations during the forming process. Secondly, unlike LSM, which requires comprehensive and uniform data for accurate modelling, DeepForge can infer the full field output from limited inputs such as surface temperature, which can be measured in-situ. Finally, the more computationally intensive level set method is less suitable for rapid adjustments to changing process conditions, highlighting DeepForge's suitability for real-time control scenarios in metal forming using MPC.

However, this architecture is not without potential drawbacks. One of the main challenges is the complexity involved in training and fine-tuning such hybrid models, especially when dealing with high-dimensional data or very complex process dynamics. \\


The integration of the DeepForge approach into a control framework, particularly for achieving desired work-piece microstructure in forging processes, is an advancement that enables in-situ online reaction on unexpected process disturbances.

DeepForge, by its design, leverages surface surface temperatures as a primary input for predicting key microstructural variables such as temperature distribution, grain size, recrystallization, and deformation. The choice to base the model on these contours is grounded in practicality and efficacy. Surface temperature data, which can be accurately and non-invasively captured using infrared (IR) cameras during the forging process.

The predictive capability of DeepForge, derived from surface temperature measurements, makes it an ideal candidate for integration into a Model Predictive Control framework. It is also important to note that only measurable surface temperatures are used as input and not, for example, the surface on which a workpiece is placed. By integrating DeepForge with MPC, the control framework can utilize predictive insights to make real-time adjustments in the process, aiming to achieve the desired microstructural properties in the work-piece. Thus, as the forging process progresses, the MPC can update its predictions and control actions based on the latest available data. This aspect is particularly advantageous in handling the dynamic nature of forging processes.

Furthermore, the time taken for optimisation within the MPC framework is an important factor. It it noted that the optimisation for the remaining parameters in the MPC framework typically takes between one to two seconds. This relatively short duration is advantageous, as it can be effectively subtracted from the wait times in the forging process, which are generally much longer, often around 20 to 30 seconds. This alignment of the optimisation time with the natural wait times in the process ensures that the implementation of MPC does not introduce significant additional delays, thus enabling real time control and  optimisation of the process parameters.

However, this aspect also highlights a limitation of the MPC application in scenarios where the process requires much shorter wait times. If the wait times in a given process are less than the time required for optimisation, which is around one to two seconds, the use of MPC might become impractical. In such cases, the time taken for optimisation by MPC could introduce delays that disrupt the process flow, negating the benefits of real-time control and adaptation. \\


Using the temperature within the developed DeepForge model within a Model Predictive Control framework enables for adaptive strategies that can influence the microstructure of a workpiece as shown in Fig \ref{fig:mpcres}.

In metal forging, the impact of the temperature variations is a critical factor to consider due to its significant influence on the metal's properties. Variations in starting temperature, often resulting from uneven heating, differences in material composition, or environmental factors, can markedly affect the metal's ductility, strength, and yield stress. For instance, metals heated to higher starting temperatures tend to be more ductile and easier to form. However, this elevated temperature can also induce grain growth, which might compromise the material's strength. On the other hand, lower starting temperatures may reduce ductility, making the metal more challenging to form, but this can also result in increased strength due to the finer grain structure that often accompanies lower temperature processing (see \citet{switzner2010effect}).

It involves dynamically adjusting various process parameters, such as heating profiles, cooling rates, and deformation speeds, to optimize the forging outcome based on the initial temperature of the workpiece. For example, when working with a workpiece that starts at a lower temperature, the process might be modified online and in-situ to apply to extend the wait time between consecutive strokes to achieve finer grainsize structure (see Figure \ref{fig:res3}). This can enhance its mechanical properties, such as strength and toughness. Finer grain sizes are generally associated with higher strength and toughness due to the increased number of grain boundaries, which impede dislocation movement, a key mechanism in metal deformation. \cite{li2007dependence, balasubramanian2016strength} \\


As far as the comparison with the state-of-the-art is concerned, this study represents a noteworthy advancement over traditional methods in the field of microstructure control in metal forming by leveraging a unique integration of machine learning architecture with Model Predictive Control (MPC) that can be used in a real-world, real-time scenario. This advancement is made possible because DeepForge operates on directly measurable inputs, such as the surface temperature of a workpiece, which can be measured in-situ, for instance, with an IR camera. Additionally, it significantly outperforms traditional finite element (FE) simulations in speed, completing simulations in approximately 4 milliseconds per stroke compared to the 20 seconds per stroke required by FE simulations. DeepForge's ability to accurately predict six different microstructural parameters based on surface temperature alone, with a margin of error of less than 1\,\% for each, is unprecedented in the state-of-the-art. Hence, it functions as a soft-sensor, enabling the prediction of non-measurable parameters using measurable ones. However, a limitation of DeepForge is that it has been trained exclusively on simulated data, meaning its accuracy is limited to that of the simulations and its precision in predicting the actual microstructure of a workpiece. Therefore, enhancements can be made through refining the simulation as well as by integrating experimental data into the dataset. Lastly, DeepForge was trained using a single billet geometry. For future development, expanding the range of geometries used in training would be advantageous to examine how DeepForge adapts to and predicts the microstructural outcomes for a variety of shapes and sizes.

\section{Summary and Conclusions}
\label{sec:summ_conc}

This paper presents a novel approach in hot closed-die forging that integrates Model Predictive Control with an advanced machine learning architecture, DeepForge, to predict and control workpiece microstructure. Furthermore, using surface temperature measurements, the feasibility and effectiveness of the MPC framework to optimise microstructure during forging, specifically grain size, by adjusting the wait time between two consecutive strokes is explored. \\

The key contributions and findings of this work are as follows:

\begin{itemize}
    \item Effective Prediction of Microstructure: The DeepForge model successfully predicts microstrucutre using surface temperature data, with the precision of 0.4$\pm$0.3\,\%.
    \item Real-Time Process Control: The integration of DeepForge with MPC enabled real-time adaptation to process disturbances, leading to the precise attainment of desired microstructural properties in the forged workpieces.
    \item Experimental Validation and Practical Application: The results of the study have been validated by experimental trials, demonstrating the practical applicability of this approach in improving the final quality of the parts produced by the closed-die forging process.
\end{itemize}

Moreover, the future work can be summarized under the following points:

\begin{itemize}
\item Deployment of DeepForge on various geometries.
\item Development of 3D version of the FE-based simulation.
\item Implementation of DeepForge within real-time forging process while utilizing a surface temperature measurement as input.
\end{itemize}

\section*{Declaration of Competing Interest}

The authors declare that they have no known competing financial interests or personal relationships that could have appeared to influence the work reported in this paper. 

\section*{Data availability}

Data will be made available on request.

\end{document}